\documentclass[10pt,twocolumn,letterpaper]{article}

\usepackage{cvpr}              

\usepackage{booktabs}
\usepackage{graphicx} %
\usepackage[dvipsnames]{xcolor}
\usepackage{algorithm}
\usepackage{mathtools}
\usepackage{svg}
\usepackage{algpseudocode}
\usepackage[ruled,vlined,algo2e]{algorithm2e}
\usepackage{marvosym}
\usepackage{multirow}
\usepackage{color}
\usepackage{colortbl}
\usepackage[font=small,skip=0pt]{caption}
\usepackage{float}
\usepackage{fontawesome}

\newcommand{\coolname}{\textit{PointINS}}

\newcommand{\PAR}[1]{\vskip2pt \noindent{\bf #1}}
\newcolumntype{?}{!{\vrule width 1pt}}

\definecolor{cvprblue}{rgb}{0.21,0.49,0.74}
\usepackage[pagebackref,breaklinks,colorlinks,allcolors=cvprblue]{hyperref}

\def\paperID{}
\def\confName{CVPR}
\def\confYear{2026}

\title{Towards Foundation Models for 3D Scene Understanding: Instance-Aware Self-Supervised Learning for Point Clouds}

\author{Bin Yang$^{1,2}$\thanks{Equal contribution.}, Mohamed Abdelsamad$^{1,3\star}$, Miao Zhang$^{1}$, Alexandru Paul Condurache$^{1,2}$\\
$^1$Bosch Center for AI, Stuttgart, Germany \\\
$^2$Institute for Neuro- and Bioinformatics, University of Lübeck, Lübeck, Germany\\\
$^3$Robert Learning Lab, University of Freiburg, Freiburg, Germany\\\
{\tt\small \{Bin.Yang3,Mohamed.Abdelsamad,Miao.Zhang5,AlexandruPaul.Condurache\}@de.bosch.com}
}

\begin{document}
\maketitle

\begin{abstract}
Recent advances in self-supervised learning (SSL) for point clouds have substantially improved 3D scene understanding without human annotations. Existing approaches emphasize semantic awareness by enforcing feature consistency across augmented views or by masked scene modeling. However, the resulting representations transfer poorly to instance localization, and often require full finetuning for strong performance. Instance awareness is a fundamental component of 3D perception, thus bridging this gap is crucial for progressing toward true 3D foundation models that support all downstream tasks on 3D data. In this work, we introduce \textbf{\coolname{}}, an instance-oriented self-supervised framework that enriches point cloud representations through geometry-aware learning. \coolname{} employs an orthogonal offset branch to jointly learn high-level semantic understanding and geometric reasoning, yielding instance awareness. We identify two consistent properties essential for robust instance localization and formulate them as complementary regularization strategies, Offset Distribution Regularization (ODR), which aligns predicted offsets with empirically observed geometric priors, and Spatial Clustering Regularization (SCR), which enforces local coherence by regularizing offsets with pseudo-instance masks. Through extensive experiments across five datasets, \coolname{} achieves on average +3.5\% mAP improvement for indoor instance segmentation and +4.1\% PQ gain for outdoor panoptic segmentation, paving the way for scalable 3D foundation models.

\end{abstract}    
\section{Introduction}
\label{sec:intro}
\begin{figure}[t]
  \centering
   \includegraphics[width=\linewidth]{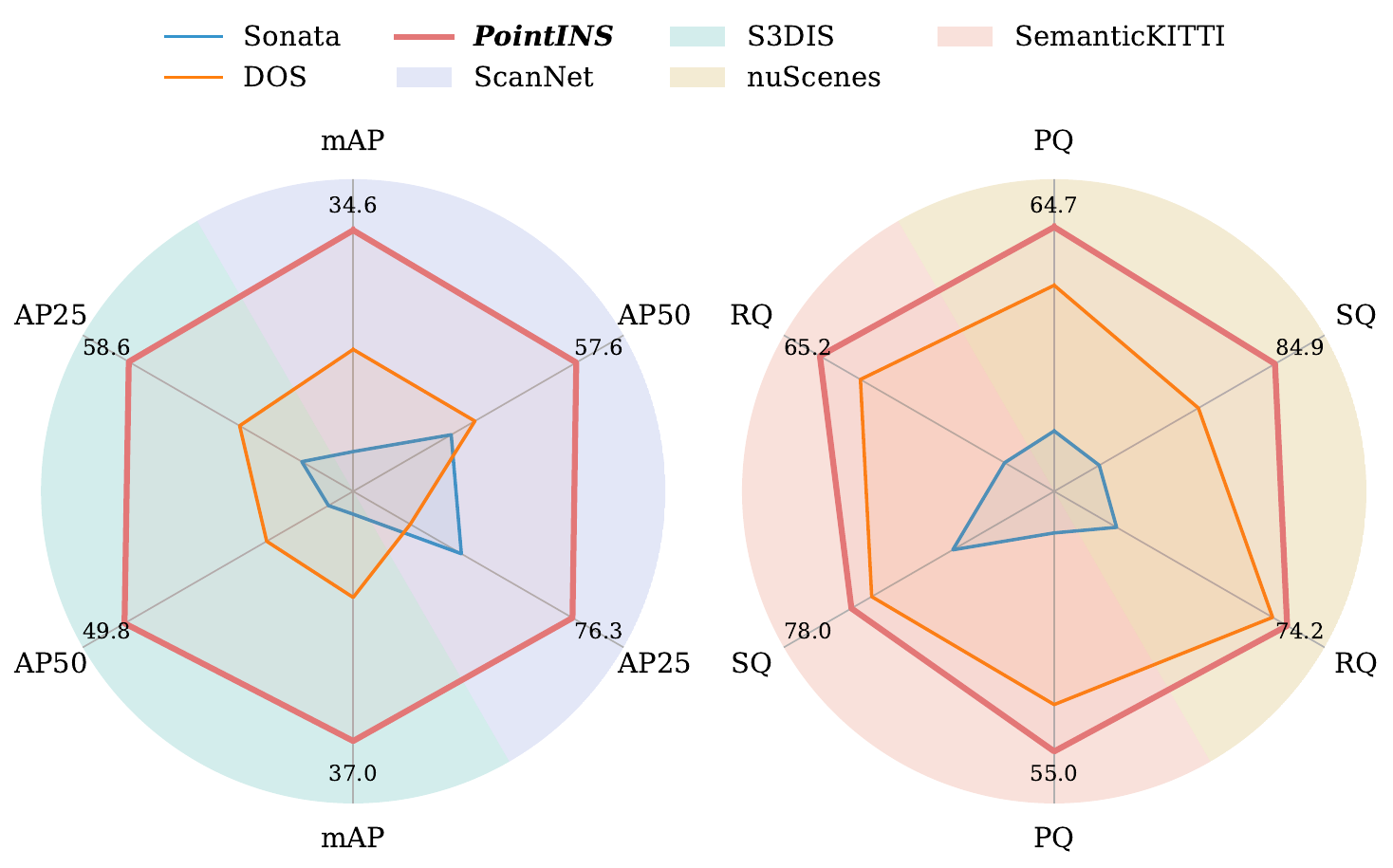}
    \caption{We achieve superior results over state-of-the-art self-supervised approaches (DOS~\cite{abdelsamad2026dos} and Sonata~\cite{wu2025sonata}) on both indoor instance (left) and outdoor panoptic segmentation (right).}
   \label{fig:intro}
   \vspace{-3mm}
\end{figure}

Self-supervised learning (SSL) has achieved great success in 2D visual representation learning~\cite{caron2021dino, he2022masked, xie2022simmim, assran2023jepa} over the past decade. It enables models to extract powerful features from large-scale unlabeled data that generalize well across diverse downstream tasks. Extending this paradigm to 3D scene understanding holds significant potential, especially in domains such as autonomous driving and robotics, where large amounts of 3D data are available but manual annotation remains costly and labor-intensive~\cite{coors2019Nova, kong2023lasermix, unal2022scribble, lowens2024unsupervised}.

\begin{figure}[b]
\centering
\vspace{-2mm}
\includegraphics[width=0.85\columnwidth]{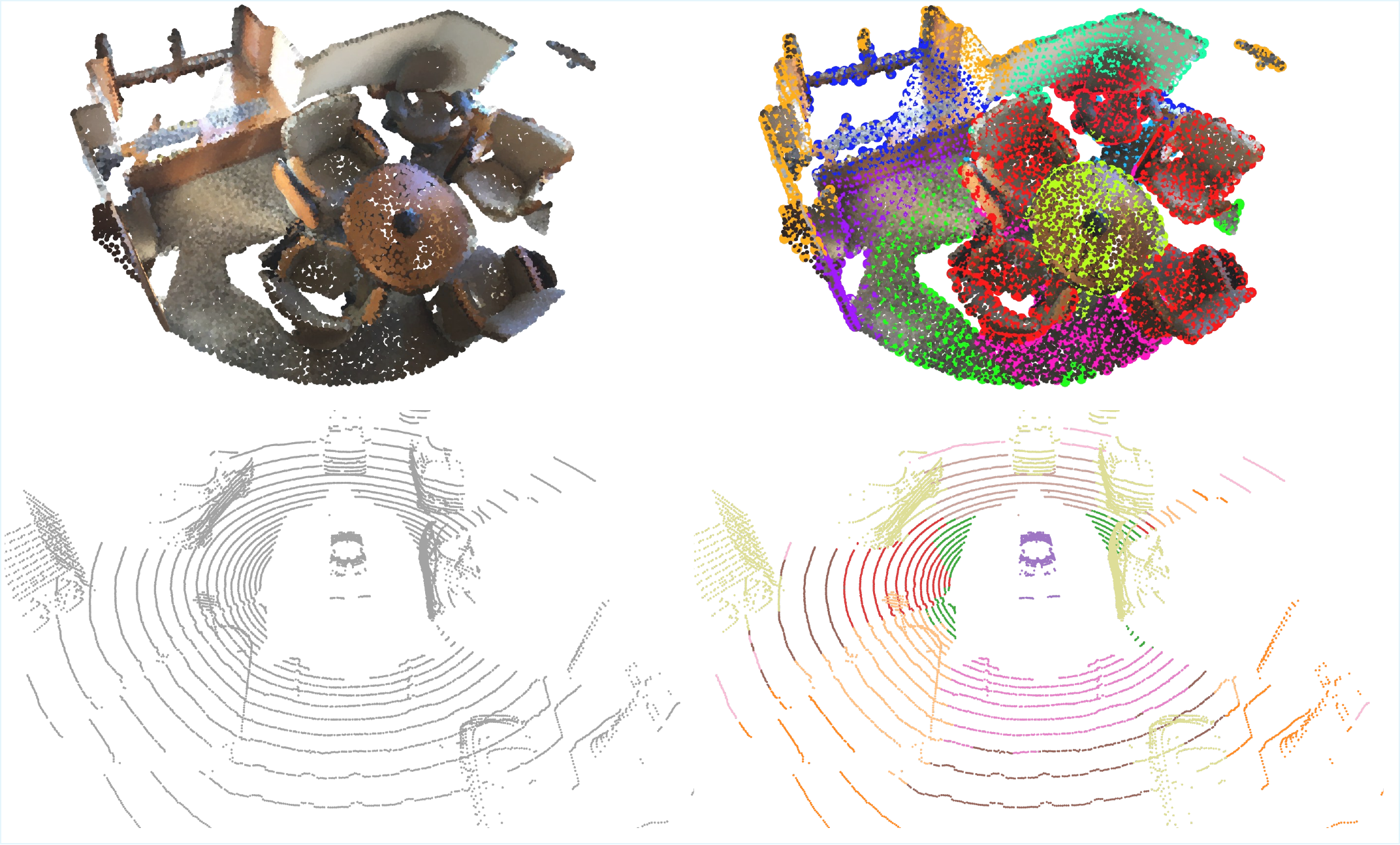}
\vspace{1mm}
\caption{Example of an indoor/outdoor scene and K-means clustering over their point features extracted from a self-supervised pre-trained model~\cite{wu2025sonata}.}
\label{fig:kmeans_example}
\end{figure}

Recent works have demonstrated promising progress in 3D SSL by applying contrastive~\cite{nunes2022segcontrast, nisar2025psa, xie2020pointcontrast} and masked modeling~\cite{wu2025sonata, hermosilla2025maskedscenemodeling, abdelsamad2026dos}. These methods achieve impressive performance on semantic segmentation benchmarks. Their success largely stems from enforcing multi-view consistency, where models are trained to produce aligned embeddings across independently augmented views of the same point cloud~\cite{wu2025sonata, caron2021dino, hermosilla2025maskedscenemodeling}. This objective effectively captures global semantic structure (see Fig.~\ref{fig:kmeans_example}). Nevertheless, such semantic-driven objectives inherently overlook instance awareness, an ability of a representation to preserve fine-grained geometric relationships. Consequently, these approaches underperform on instance-oriented tasks such as instance and panoptic segmentation. Closing this gap is essential for advancing unified 3D foundation models that can generalize across diverse tasks and domains. 

A key challenge lies in balancing semantic invariance with the geometric sensitivity needed for learning instance awareness. Prior works~\cite{wu2025sonata, hermosilla2025maskedscenemodeling} have identified that point cloud SSL can collapse to trivial cues like normals or poses of points, thus, most methods enforce strong invariance to avoid such geometric shortcuts. While we acknowledge this concern, we argue that the geometric proximity required for instance-aware learning represents a high-level relational property rather than a low-level geometric cue. We refer to this capability as \textit{geometric reasoning}. Beyond learning consistency for semantics, SSL features should encode where a point should direct, ideally toward the centroid of the instance it belongs to. Our hypothesis aligns with supervised instance and panoptic segmentation frameworks~\cite{jiang2020pointgroup, zhou2021panoptic, li2023center, li2022panoptic}, where semantic categorization guides subsequent geometric grouping into class-agnostic instances. These architectures typically use a shared backbone with parallel semantic and offset branches: semantics restrict the candidate regions for instance boundaries, while offsets refine the spatial separation. By reinforcing each other, they jointly enhance the model's overall understanding of 3D scenes.

Building upon this insight, we propose \coolname{}, the first self-supervised framework explicitly designed to learn both semantic consistency and geometric reasoning. Our framework augments the SSL framework with an additional instance-localization branch that predicts point-wise geometric offsets toward underlying instance centers. 

However, predicting the offsets without labels remains challenging. Without ground-truth guidance, the model struggles to infer meaningful geometric relationships, and may even collapse to trivial solutions~\cite{caron2021dino, assran2023jepa}. At the same time, we observe that offsets inherently exhibit several stable statistical and structural properties that can be exploited to guide and regularize the learning process. To this end, we introduce two complementary regularization strategies: \textbf{Offset Distribution Regularization (ODR)} provides a global constraint by aligning predicted offsets with empirical geometric priors observed in real scenes, while \textbf{Spatial Clustering Regularization (SCR)} introduces a local constraint by enforcing consistent centroid alignment among points through pseudo-instance masks derived from the model’s semantic understanding. Together, these components provide both global distributional alignment and local geometric coherence to enable robust instance-aware representation learning. 

We evaluate \coolname{} extensively on both indoor and outdoor scene datasets, covering semantic, instance, and panoptic segmentation under linear probing, decoder probing, and finetuning protocols. Our method achieves state-of-the-art performance among self-supervised approaches, yielding 2.5–4.6\% mAP gains on indoor instance segmentation and 3.4–4.8\% PQ gains on outdoor panoptic segmentation. To summarize, this work contains the following key contributions:
\begin{itemize}[topsep=-5px,partopsep=0px]
    \item We introduce a novel self-supervised training framework for point clouds that jointly learns semantic consistency and geometric reasoning.
    \item We identify two consistent properties essential for robust instance awareness and introduce complementary regularization strategies that prevent model collapse while enabling effective representation learning.
    \item Our method surpasses the state-of-the-art methods by a clear margin on indoor instance and outdoor panoptic segmentation, representing a step toward unified 3D foundation models for holistic scene understanding.
\end{itemize}

\section{Related Works}
\label{sec:related_works}
\subsection{3D Instance \& Panoptic Segmentation}
3D instance segmentation methods typically rely on dense supervision to learn point-wise offsets or embeddings that facilitate object grouping. Early approaches~\cite{jiang2020pointgroup, chen2021hais, vu2022softgroup} predict offset vectors pointing toward object centers and cluster points using mean-shift algorithms or connected components. Panoptic segmentation extends this paradigm by jointly predicting semantic labels and instance identities for every point in the scene~\cite{fong2022panoptic, mei2022waymo}. Many of these methods adopt a dual-branch design, where semantic and instance segmentation are learned simultaneously with a shared backbone. While highly effective, they depend on a large amount of instance and semantic annotations for fully supervised training. In this work, we seek to learn both semantic and instance-aware representations directly from unlabeled point clouds for enhancing performance on diverse segmentation tasks.
 
\subsection{Point Cloud Self-supervised Learning}
Early self-supervised learning (SSL) approaches~\cite{wu2023msc, xie2020pointcontrast, yang2023gdmae} adopt primarily 3D sparse convolutions with U-Net-like architectures~\cite{choy2019minkowsiki, cciccek20163dunet}. These works fall broadly into two categories: contrastive learning~\cite{nunes2022segcontrast, zhang2021depthcontrast, xie2020pointcontrast, wu2023msc, nisar2025psa} and occupancy reconstruction~\cite{abdelsamad2025nomae, yang2023gdmae, tian2023geomae,min2023occupancy, krispel2024maeli}. In contrastive frameworks, points or patches from the same scene are treated as positive pairs and encouraged to produce similar embeddings. However, these objectives lack spatial cues to distinguish nearby objects belonging to the same semantic class. Occupancy reconstruction approaches learn to reconstruct missing geometry from visible patches. Although this enhances geometric understanding, the optimization primarily targets local completeness rather than instance awareness. Moreover, both paradigms typically require full model finetuning for downstream tasks, which is computationally expensive for large-scale 3D data. Recently, transformer-based backbones have demonstrated strong generalization and scalability across different 3D perceptual tasks~\cite{wu2024ptv3, zhao2021point, he2022voxelset, yang2026flares, yang2024tulip}. This advancement motivates newer SSL methods to adopt them~\cite{wu2025sonata}. These methods show strong performance in semantic segmentation under linear probing setting by enforcing feature consistency across independently augmented views. However, this strong focus promotes semantic-compact but geometry-entangled representations, which suppresses the intra-class variation needed for instance-level perception. 
\section{Method}
\label{sec:method}
In this section, we detail the technical components of our approach. We first introduce the overall teacher–student architecture, which employs a prototype-based self-distillation mechanism as the semantic branch for learning category-level understanding. We then describe the offset branch, where the model learns to reason about geometric relationships among points. Finally, we present two complementary regularization strategies that stabilize training and strengthen the geometric reasoning capability essential for offset-based instance-aware learning.

\subsection{Preliminary}
\begin{figure*}[t]
\vspace{-4mm}
\centering
    \includegraphics[width=\linewidth]{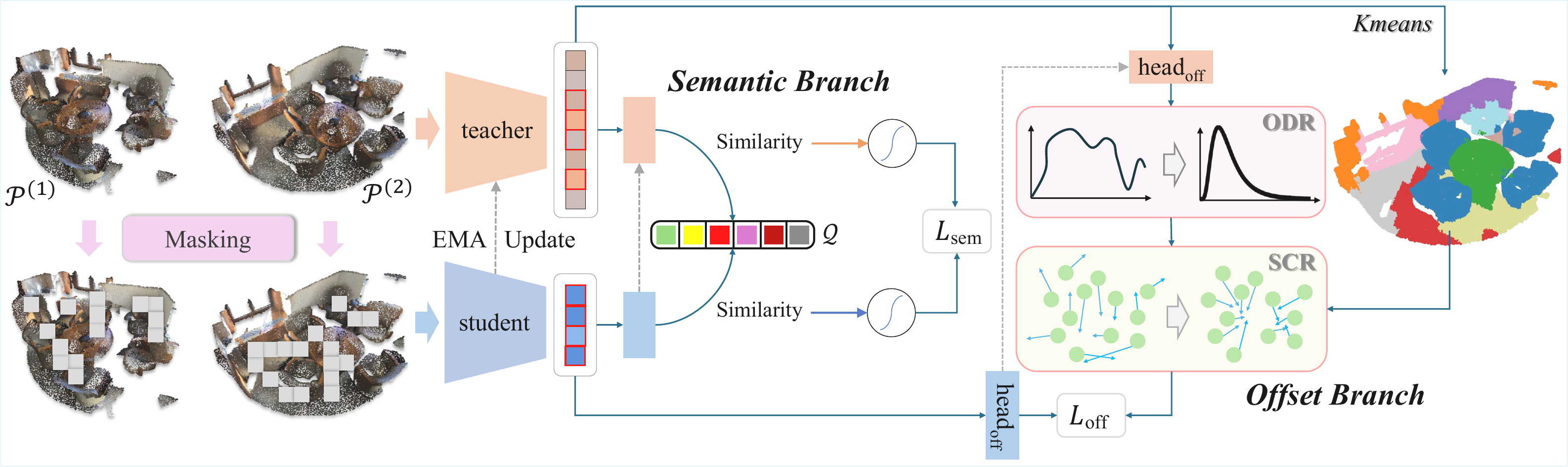}
\caption{Overview of \coolname{}: A point cloud is augmented to two independent views and they are randomly masked. The teacher processes the full input and the student receives only visible points. Both networks share the same architecture. In the semantic branch, features are computed similarity with prototypes $\mathcal{Q}$. A KL-divergence loss $L_{\text{sem}}$ is then applied for distillation. In the offset branch, an offset head maps features into 3D offset vectors. Teacher offsets are first regularized by ODR to align with empirically observed geometric priors. Next, segments obtained from K-means-clustering are used to extract pseudo-instance masks. Those masks help to enhance instance awareness by regularizing local coherence of points. Finally, an offset loss $L_{\text{off}}$ is computed as the second distillation.  }
\label{fig:overview}
\end{figure*}
%
Our framework builds upon a widely adopted teacher-student self-distillation paradigm in recent 3D SSL methods with hierarchical decoder-free architectures~\cite{wu2025sonata, abdelsamad2026dos}. As shown in Fig.~\ref{fig:overview}, a point cloud $\mathcal{P}={(x_i,f_i)}_{i=1}^N$, where $x_i$ denotes the coordinates and $f_i$ the feature of point $i$, is randomly augmented into two distinct views, $\mathcal{P}^{(1)}$ and $\mathcal{P}^{(2)}$. Each view undergoes independent spatial and photometric augmentations. Then, we randomly mask a subset of points to create a visible subset $\mathcal{P}_v$ as inputs for the student network, while the teacher receives the full point cloud $\mathcal{P}$. Both networks share identical architectures, and the teacher’s parameters are updated by the exponential moving average (EMA) of the student’s. 

For the semantic branch, we adopt a prototype-based clustering mechanism~\cite{ caron2021dino}. Both teacher and student encode augmented views into point-wise embeddings, which are projected onto a set of learnable category prototypes $\mathcal{Q}={q_k}_{k=1}^K$ by computing the similarity between embeddings and prototypes. Then, the similarity is transformed into soft assignments via a temperature-scaled softmax. The student is trained to align its distributions with those of the teacher via a Kullback–Leibler (KL) divergence loss. As the student process only visible subset of points, we select the corresponding tokens in teacher for computing the loss. To enforce the semantic consistency across views, the loss is computed second time for cross-view distillation ($\mathcal{P}^{(2)} \rightarrow  \mathcal{P}^{(1)}_v$ and $\mathcal{P}^{(1)} \rightarrow \mathcal{P}^{(2)}_v$). 

\subsection{Learning Offset without Labels}
To inject instance awareness into self-supervised learning, we introduce an offset branch that predicts a 3D vector for each point, which directs toward the geometric center of its underlying instance. Unlike the semantic branch, which learns view-consistent embeddings, this branch captures view-dependent geometric relationships and is therefore sensitive to spatial transformations such as rotation, flipping, and scaling. To maintain geometric consistency, we track the transformations applied during data augmentation and invert them to map the predicted offsets back to the original coordinate. 

\begin{figure}[h]
\vspace{-2mm}
\centering
    \includegraphics[width=0.75\columnwidth]{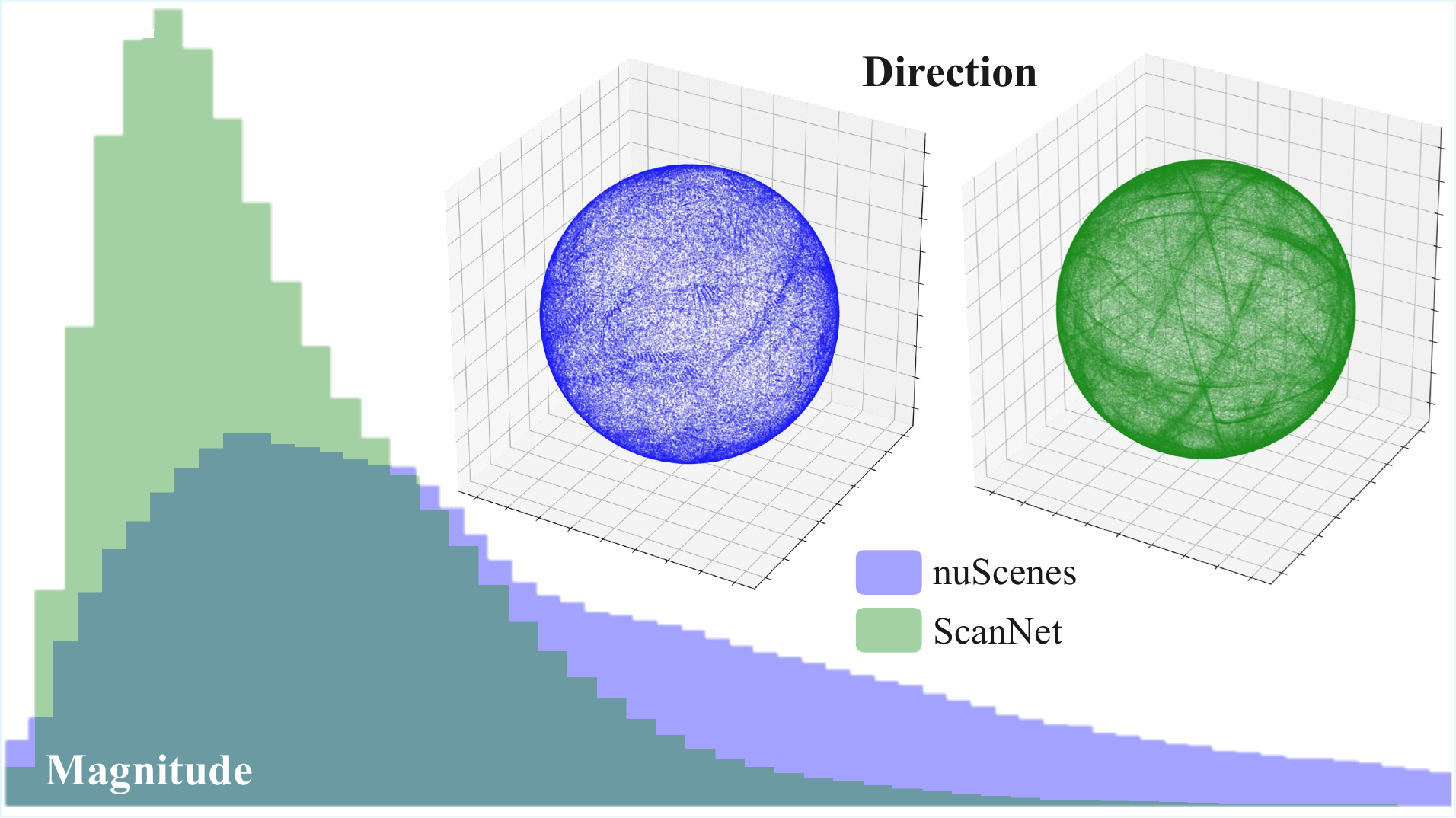}
\caption{Offset distributions of ScanNet~\cite{dai2017scannet} and nuScenes~\cite{fong2022panoptic}}
\label{fig:statistics}
\vspace{-3mm}
\end{figure}

\PAR{Offset Distribution Regularization (ODR)} Regressing offsets without supervision can lead to collapsed predictions~\cite{caron2021dino}. Therefore, directly introducing offset learning into the self-supervised stage risks unstable optimization. To mitigate this issue, we introduce Offset Distribution Regularization (ODR), which constrains the predicted offsets to match statistically grounded distributions observed in real scenes. Each offset vector $\mathcal{O} \in \mathbb{R}^3$ can be decomposed into two components: (1) the magnitude, which measures the distance to the instance centroid, and (2) the direction, a unit vector pointing toward that centroid. From our analysis of existing scene datasets, we observe two consistent trends: (1) offset magnitudes follow a stable long-tailed distribution reflecting scene layout and object scale, and (2) offset directions are approximately uniformly distributed over the unit sphere (see Fig.~\ref{fig:statistics}). These observations motivate us to adopt them as global statistical priors for regularizing the predicted offsets.

\begin{figure}[h]
\vspace{-3mm}
\centering
    \includegraphics[width=0.8\columnwidth]{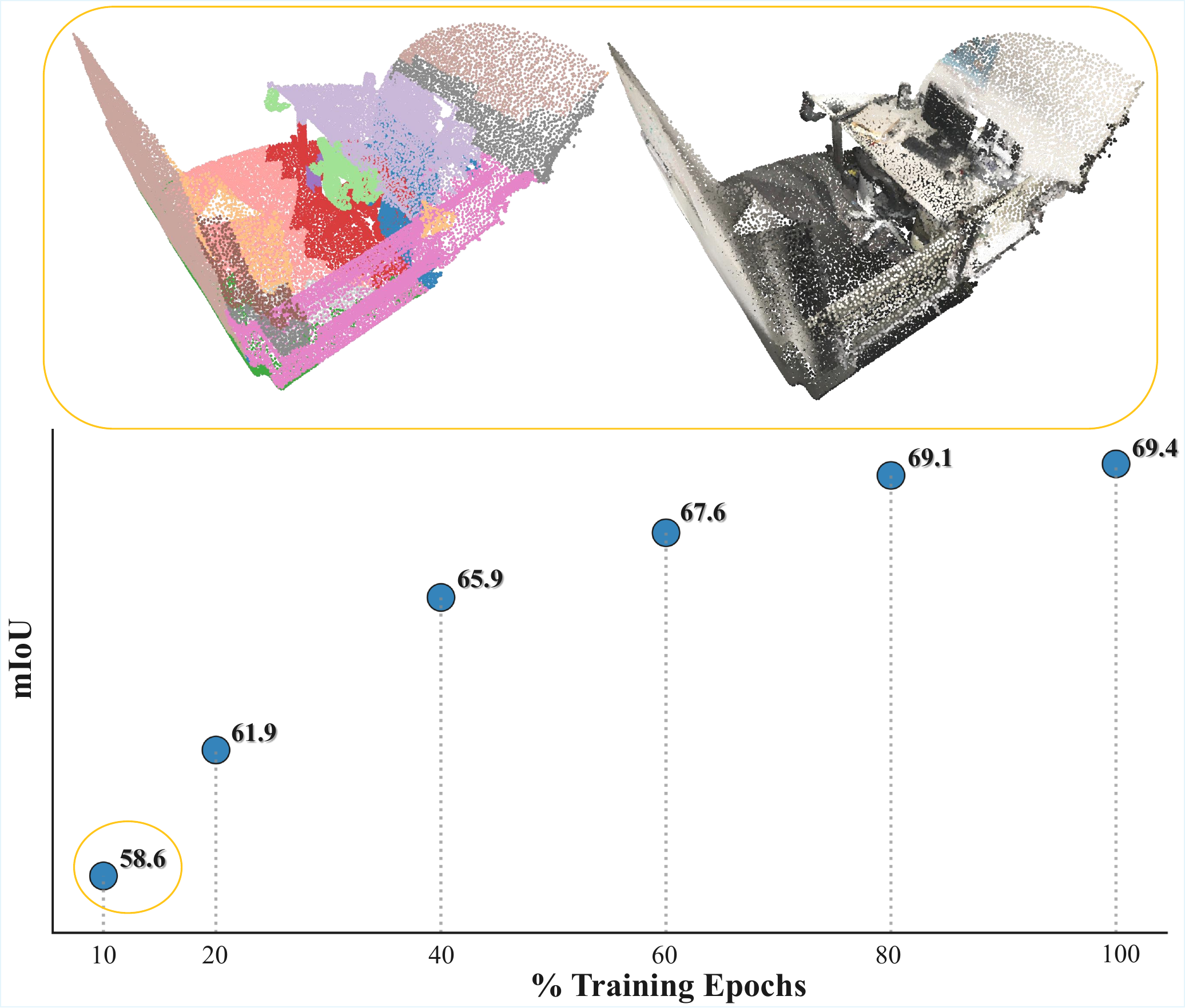}
\caption{We monitor the linear probing (LP) performance of a model pre-trained with a recent SSL method~\cite{wu2025sonata}. Remarkably, the model reaches 85\% of its final LP performance within just 10\% of the total training.}
\label{fig:kmeans_segments}
\vspace{-2mm}
\end{figure}

We achieve this using the \textbf{probabilistic integral transform (PIT)}, a non-parametric method that maps scalar samples to a target distribution while maintaining their relative order. Given the predicted offset magnitudes $\{\mathcal{M}_i\}_{i=1}^N \in \mathbb{R}$, we convert them to PIT-normalized values as follows. Let $\pi(i)$ denote the rank of $\mathcal{M}_i$ in ascending order:

\[
\pi(i) = \mathrm{rank}(\mathcal{M}_i), \quad 1 \leq \pi(i) \leq N,
\]
which we convert into probability levels via
\[
u_i = \frac{\pi(i) - 0.5}{N}, \quad u_i \in (0, 1).
\]
We then transform these into target-aligned magnitudes by applying the inverse cumulative distribution function (CDF) $F^{-1}$ of the empirical distribution observed in real scenes:
\[
\tilde{\mathcal{M}}_i = F^{-1}(u_i).
\]
This procedure strictly preserves rank order (i.e., $\mathcal{M}_i < \mathcal{M}_j \iff \tilde{\mathcal{M}}_i < \tilde{\mathcal{M}}_j$) while aligning the predicted magnitudes to the desired long-tailed distribution. Since PIT operates on scalar values, we apply it independently to each coordinate of direction $\{\mathcal{D}_i\}_{i=1}^N \in \mathbb{R}^3$ to match the uniform distribution on each axis. The complete offset after PIT-normalization is then reconstructed as
\[
\tilde{\mathcal{O}}_i = \tilde{\mathcal{M}}_i \cdot \tilde{\mathcal{D}}_i.
\]
This regularization encourages geometrically plausible offset predictions while preventing model collapse, thereby stabilizing training without labels.
\PAR{Spatial Clustering Regularization (SCR)} While effective, ODR alone cannot reason about local coherence for instance localization, i.e. points in the local neighborhood should predict offsets converging toward a common centroid when they inherently belong to the same instance. Without additional constraint, predicted offsets may remain scattered irregularly. To address the limitation, we introduce another regularization technique, so-called Spatial Clustering Regularization (SCR). In our empirical study, we found that features learned by recent SSL frameworks naturally exhibit strong semantic awareness even in the early training stage (see Fig.~\ref{fig:kmeans_segments}). Inspired by this observation, we use these features to generate pseudo-instance masks that guide the model toward instance-aware learning at the pre-training stage. 

Specifically, we begin by applying K-means clustering to the point-wise features $\mathbf{F} = \{ f_i \}_{i=1}^N$ extracted by the teacher network and obtain $K$ class-wise segments:
\begin{equation}
    \mathcal{S} = \{ S_1, S_2, \dots, S_K \} 
\end{equation}
For each point, we compute its predicted centroid by adding the PIT-normalized offset to its 3D coordinate:
\begin{equation}
    \hat{c}_i = x_i + \tilde{\mathcal{O}}_i,
\end{equation}
Within each segment $S_k$, we further refine the structure by building a local $k$-nearest-neighbor graph over these predicted centroids and applying Breadth-First Search (BFS) to break the cluster into multiple connected components:
\begin{equation}
    \mathcal{I}_k = \{ I_{k,1}, I_{k,2}, \dots \}, \quad I_{k,j} \subseteq S_k.
\end{equation}
These spatially coherent components serve as pseudo-instances. For each pseudo-instance $I_{k,j}$, we update the centroid as:
\begin{equation}
    \bar{c}_{k,j} = \frac{1}{|I_{k,j}|} \sum_{i \in I_{k,j}} x_i,
\end{equation}
 The updated centroid $\bar{c}_{k,j}$ then becomes the new supervision target for each point, and the corresponding target offset is recalculated as:
\begin{equation}
    \mathcal{O}_i^\ast = \bar{c}_{k,j} - x_i,
\end{equation}
This regularization reinforces local geometric consistency and further stabilize the instance-aware learning in the self-supervised setting.

\PAR{Complementarity of Two Regularization Strategies} SCR enforces local coherence by grouping spatially adjacent points into pseudo-instances, thereby complementing the global structural regularization of ODR. Conversely, ODR enhances the stability of SCR. Since SCR relies on spatial proximity for instance cutting, scattered or unstable offset predictions can lead to irregular grouping. ODR addresses this by constraining the offsets via geometric priors, which provides a stable geometric anchor that guides SCR toward consistent clustering. This complementarity between these two regularization strategies is demonstrated qualitatively in Fig.~\ref{fig:regularization_qualitative} and quantitatively in Sec.~\ref{sec:experiment}.
\PAR{Offset Self-distillation} After deriving refined centroids through ODR and SCR, we use them to supervise the student network via self-distillation. Following the practice in instance segmentation methods~\cite{jiang2020pointgroup, zhou2021panoptic}, the offset loss consists of an $\ell_1$ loss that penalizes deviations in offset magnitude, and a cosine similarity loss that aligns the direction of predicted offsets with the target. Formally, the offset loss is:
\begin{equation}
\mathcal{L}_{\text{offset}} = \frac{1}{N} \sum_{i=1}^{N} \left( \| {o}_i - \mathcal{O}_i^\ast \|_1 \;+\;  \left( 1 - \cos(o_i, \mathcal{O}_i^\ast) \right) \right),
\end{equation}
where $o_i$ is the predicted offset of student network and $\mathcal{O}_i^\ast$ is the updated target offset. Same as the semantic branch, a cross-view loss is applied to enforce offset consistency between two augmented views. For stable optimization, offset loss is delayed to join the training after few warm-up epochs and a loss weight $\lambda_{\text{off}}$ is integrated for balancing. The full procedure is summarized in Algo.~\ref{alg:offset_loss}.

\begin{algorithm}[t]
\caption{Pre-training procedure}
\label{alg:offset_loss}
\begin{algorithmic}[1]
\Require Full Point cloud $\mathcal{P}$ and Visible Subset $\mathcal{P}_v$ 
        Teacher $T$, Student $S$, Prototypes $\mathcal{Q}$, Projection Head $\mathcal{H}$, Offset Head $\mathcal{H}_{\text{off}}$
      \State \textbf{Student/Teacher features:}
          \Statex  $\mathbf{F}^{(1)}_{S} \gets S(\mathcal{P}_v^{(1)})$, \, $\mathbf{F}^{(2)}_{S} \gets T(\mathcal{P}_v^{(2)})$
          \Statex  $\mathbf{F}^{(1)}_{T} \gets T(\mathcal{P}^{(1)})$, \, $\mathbf{F}^{(2)}_{T} \gets T(\mathcal{P}^{(2)})$
      \State \textbf{Semantic self-distillation:}
      \Statex  $L_{\text{sem}} \gets \mathrm{KLDiv}\!\big(\sigma(\mathcal{H}(\mathbf{F}^{(1)}_{S}),\mathcal{Q}),\, \sigma(\mathcal{H}(\mathbf{F}^{(1)}_{T}), \mathcal{Q})\big)$, 
          \Statex where $\sigma(\cdot)$ is softmax operator.
      \State \textbf{Offset distribution regularization (ODR):}
      \Statex $\hat{\mathcal{O}}_{T} \gets \mathcal{H}_{\text{off}}(\mathbf{F}^{(1)}_{T})$; \, $\tilde{\mathcal{O}}_{T} \gets \mathrm{ODR}(\hat{\mathcal{O}}_{T})$
      \Statex  $\hat{c}_i \gets x_i + \tilde{\mathcal{O}}_{T,i}$
      \State \textbf{Spatial clustering regularization (SCR):}
      \Statex  $\{S_k\} \gets \mathrm{KMeans}(\mathbf{F}^{(1)}_{T})$
       \Statex \text{\textbf{for} each segment $S_k$ \textbf{do:}} \quad $\{I_{k,j}\} \gets \mathrm{BFS}(\{\hat{c}_i\}_{i \in S_k})$
        \State \textbf{Update the centroids}
        \Statex $\bar{c}_{k,j} \gets \frac{1}{|I_{k,j}|} \sum_{i \in I_{k,j}} x_i$
        \State \textbf{Get the new teacher offsets}
        \Statex $\mathcal{O}^\ast_i \gets \bar{c}_{k,j} - x_i$ \textbf{ for all } $i \in I_{k,j}$
      \State \textbf{Offset self-distillation:}
      \Statex \quad $o_{S} \gets  \mathcal{H}_{\text{off}}(\mathbf{F}^{(1)}_{S})$
      \Statex \quad $L_{\text{off}}=\mathcal{L}_{\text{offset}}(o_{S}, \mathcal{O}^\ast)$
      \State \textbf{Optimization: } 
      \Statex $L \gets L_{\text{sem}} + \lambda_{\text{off}} L_{\text{off}}$
       \Statex Update $S$ by backprop and $T$ by EMA
\end{algorithmic}
\end{algorithm}
\PAR{Design Rationale} Our key insight is to formulate instance-aware learning as a regularized self-distillation problem, where ODR and SCR are applied on the teacher side to progressively refine offset predictions toward geometrically valid solutions, providing a stable supervisory signal without interfering with the student's representation learning. We choose point-wise offsets as the new target for their simplicity, enabling seamless integration into existing SSL frameworks with minimal additional model capacity.

\section{Experiment}
\label{sec:experiment}
\begin{table*}[h]
\vspace{-3mm}
\centering
\resizebox{0.85\linewidth}{!}{
\begin{tabular}{l?c|ccc?c|ccc?c|ccc}
\toprule
\multirow{3}{*}{\textbf{Method}} & 
\multicolumn{4}{c?}{\textbf{ScanNet val}} & 
\multicolumn{4}{c?}{\textbf{ScanNet200 val}} & \multicolumn{4}{c}{\textbf{S3DIS Area5}} \\
\cmidrule(lr){2-5} \cmidrule(lr){6-9} \cmidrule(lr){10-13} 
&SemSeg.& \multicolumn{3}{c?}{InsSeg.} &SemSeg.& \multicolumn{3}{c?}{InsSeg.}&SemSeg.& \multicolumn{3}{c}{InsSeg.}\\
\cmidrule(lr){2-2} \cmidrule(lr){3-5} \cmidrule(lr){6-6} \cmidrule(lr){7-9} \cmidrule(lr){10-10} \cmidrule(lr){11-13}
& mIoU & mAP & AP$_{50}$ & AP$_{25}$ & mIoU & mAP & AP$_{50}$ & AP$_{25}$ & mIoU & mAP & AP$_{50}$ & AP$_{25}$ \\ 
\midrule
\rowcolor{gray!7} PTv3 (sup.) & 77.6 & 40.9 & 61.7 & 77.5 & 35.3& 24.0& 34.1& 40.8 & 73.4 & 40.2 &52.1 &61.2\\
\midrule
SegContrast~\cite{nunes2022segcontrast}(lin.)  & 38.4 & 6.4 & 13.7 & 30.6 & 12.4& 1.8& 4.7& 10.2& -& -& -&- \\
PSA~\cite{nisar2025psa} (lin.) & 42.9 & 9.7 & 20.4 & 41.9 & 13.9& 2.4& 5.3& 11.4& -& -& -&- \\
NOMAE~\cite{abdelsamad2025nomae}  (lin.)& 47.5  & 9.5 & 20.0 & 42.0 & 14.7& 2.9& 6.7& 12.7& -& -& -&- \\
Sonata~\cite{wu2025sonata} (lin.) & 67.4  & 25.0 & 46.1 & 64.6 & 26.9& 8.7& 17.5& 25.2& 69.3 & 24.2 &33.8 &47.5\\
DOS~\cite{abdelsamad2026dos} (lin.) &\textbf{72.8}  & \underline{28.7} & \underline{49.8} & \underline{68.7} & \underline{29.1}& \underline{10.9}& \underline{20.6}&\underline{27.9} & \underline{70.6} & \underline{28.6} & \underline{41.0}& \underline{52.3} \\
\rowcolor{green!8} \coolname{} (lin.) & \underline{72.4}   & \textbf{32.1} & \textbf{55.2} & \textbf{73.6} & \textbf{29.6}&\textbf{13.4} &\textbf{24.9} &\textbf{35.8} & \textbf{71.0} & \textbf{33.2}& \textbf{45.3}&\textbf{59.4}\\
\bottomrule
SegContrast~\cite{nunes2022segcontrast}(dec.)  & 67.1 & 30.5 & 50.3 & 66.9 &24.8 &10.5 &18.0 &25.5 & -& -& -&- \\
PSA~\cite{nisar2025psa} (dec.) &67.1  & 28.9 & 47.3 & 64.5 & 24.3& 10.9& 18.0&25.5 & -& -& -&- \\
NOMAE~\cite{abdelsamad2025nomae} (dec.)& 68.0 & 28.5 & 49.3 & 68.0 & 30.5& 15.8&24.8 &31.8 & -& -& -&- \\
Sonata~\cite{wu2025sonata} (dec.) & 75.5 & 37.1 & 57.8 & 74.2 & 31.6& 17.9& 27.8& 34.8& 71.8 &36.8& \underline{48.3}& \underline{60.6} \\
DOS~\cite{abdelsamad2026dos} (dec.)  & \textbf{76.9} & \underline{38.9} & \underline{60.2} & \underline{75.5} & \underline{33.7} & \underline{18.8}& \underline{28.5}& \underline{36.2}&  \textbf{73.0}& \underline{37.2} & 47.4 &60.1\\
\rowcolor{green!8} \coolname{} (dec.) & \underline{76.7} &  \textbf{40.2}& \textbf{62.5} & \textbf{77.2} & \textbf{33.9} & \textbf{21.3} & \textbf{32.2}  & \textbf{38.3}& \underline{72.9} & \textbf{39.1}& \textbf{51.7}& \textbf{61.7}\\
\bottomrule
SegContrast~\cite{nunes2022segcontrast}(fin.)  & 75.5  & 39.8 & 60.5 & 75.2 & 33.8& 19.6 &29.2 &36.5 & -& -& -&- \\
PSA~\cite{nisar2025psa} (fin.) & 76.2 & 38.8 & 59.7 & 74.3 & 33.5&19.4 &28.0 &35.6 & -& -& -&- \\
NOMAE~\cite{abdelsamad2025nomae}  (fin.)& 75.3  & 37.0 & 59.5 & 76.4 & 33.0 &19.3 &28.4 & 36.0& -& -& -&- \\
Sonata~\cite{wu2025sonata} (fin.) & 77.2  & 39.5 & 61.1 & 76.7 & 34.4&20.0 &28.5 & 35.8& 73.4 & \underline{41.3} & \underline{53.3}& \underline{62.2}\\
DOS~\cite{abdelsamad2026dos} (fin.) & \underline{78.7}  & \underline{40.5} & \underline{62.0} & \underline{77.3} & \textbf{36.7}& \underline{22.5} & \underline{33.5} & \underline{38.6} & \textbf{74.2} & 40.4 & 52.0 & 60.3\\
\rowcolor{green!8} \coolname{} (fin.) & \textbf{79.0}  & \textbf{41.5} & \textbf{63.7} & \textbf{78.4}  & \underline{36.6} & \textbf{25.1} &\textbf{34.8} & \textbf{41.5} & \underline{73.6}& \textbf{42.9}& \textbf{57.2} & \textbf{64.1}\\
\bottomrule
\end{tabular}
}
\caption{Comparison of self-supervised methods on various indoor scene datasets: ScanNet~\cite{dai2017scannet}, ScanNet200~\cite{rozenberszki2022scannet200} and S3DIS~\cite{landrieu2018s3dis}. For fair comparison, all models deploy the same backbone of Point Transformer v3 (PTv3)~\cite{wu2024ptv3} and no additional data is used during pre-training. The \textbf{best} and \underline{second best} results in each setting are highlighted in \textbf{bold} and \underline{underline}, respectively.
}
\vspace{-3mm}
\label{tab:main_results}
\end{table*}

\PAR{Implementation Details} We build our method upon the DOS~\cite{abdelsamad2026dos} pre-training framework. The backbone follows a lightweight, decoder-free variant of Point Transformer V3 (PTv3)~\cite{wu2024ptv3}. Multi-scale features from the encoder are upsampled to a common resolution and concatenated. The offset branch consists of two MLP layers that map the concatenated features to 3D offset vectors. For downstream evaluation, we follow PointGroup~\cite{jiang2020pointgroup} for training and test under three standard protocols: linear probing, where the backbone is frozen and only a head is trained; decoder probing, where a standard decoder is trained; and full finetuning. In SCR, we use $K = 20$ clusters for the K-means step. We set the offset loss weight $\lambda_{\text{off}} = 0.25$ and apply a warm-up schedule for the first 10\% of training epochs. Additional implementation details are provided in the Appendix.
\begin{table}[h]
\centering
\resizebox{\linewidth}{!}{
\begin{tabular}{l?c|ccc?c|ccc}
\toprule
\multirow{3}{*}{\textbf{Method}} & 
\multicolumn{4}{c?}{\textbf{nuScenes val}}   & \multicolumn{4}{c}{\textbf{SemanticKITTI val}}\\
\cmidrule(lr){2-5} \cmidrule(lr){6-9} 
&SemSeg.& \multicolumn{3}{c?}{PanSeg.} &SemSeg.& \multicolumn{3}{c}{PanSeg.}\\
\cmidrule(lr){2-2} \cmidrule(lr){3-5} \cmidrule(lr){6-6} \cmidrule(lr){7-9} 
& mIoU &PQ& SQ & RQ & mIoU &PQ& SQ & RQ\\ 
\midrule
\rowcolor{gray!7} PTv3 (sup.) & 80.3 & 69.9 & 86.3 & 80.5  &69.1 & 58.2& 78.7 & 67.7 \\
\midrule
SegContrast~\cite{nunes2022segcontrast}(lin.) & 37.8& 25.4 & 69.8 & 33.4&- & -& -&-\\
PSA~\cite{nisar2025psa} (lin.) & 44.5 & 30.1& 73.9&38.6&- & -& -&-\\
NOMAE~\cite{abdelsamad2025nomae}  (lin.) & 64.7  & 45.5 & 77.0 & 56.4  & 29.8 &17.1 &54.7 &24.9\\
Sonata~\cite{wu2025sonata} (lin.) &59.2  & 50.7 & 79.8 & 61.6& 46.5&34.5 & 65.5&44.2 \\
DOS~\cite{abdelsamad2026dos} (lin.) & \underline{74.1} & \underline{57.4} & \underline{82.8} & \underline{68.5}  & \textbf{67.5} & \underline{49.6} &\underline{71.8} & \underline{60.9}\\
\rowcolor{green!8} \coolname{} (lin.) & \textbf{74.4} & \textbf{62.2} & \textbf{84.5} & \textbf{72.8} & \underline{66.9} & \textbf{52.8} & \textbf{73.7} & \textbf{63.2}\\
\bottomrule
SegContrast~\cite{nunes2022segcontrast}(dec.)  & 73.1  & 60.7 & 83.9 &71.7& - & -& -&- \\
PSA~\cite{nisar2025psa} (dec.) &74.6 &62.2  & 84.1 & 73.3 &- & -& -&-\\
NOMAE~\cite{abdelsamad2025nomae} (dec.)&\textbf{80.1} & 69.0 & \underline{85.6} & \underline{79.3} &64.3 & 52.4&73.6 &62.3\\
Sonata~\cite{wu2025sonata} (dec.) & 76.8 & 66.0 &85.2  &76.8 &64.5 & 55.1& 77.5& 64.8\\
DOS~\cite{abdelsamad2026dos} (dec.)  & 79.2 & \underline{69.1} & 85.3 & 79.0  & \textbf{68.6} & \underline{56.7} &\underline{76.3} & \underline{65.2}\\
\rowcolor{green!8} \coolname{} (dec.) &  \underline{80.0} & \textbf{70.8}  & \textbf{86.6}  & \textbf{80.6}  & \underline{68.2}& \textbf{59.2}& \textbf{78.2}& \textbf{68.7}\\
\bottomrule
SegContrast~\cite{nunes2022segcontrast}(fin.)  &78.0& 69.3 & 85.9 & 80.3& - &- & -& -\\
PSA~\cite{nisar2025psa} (fin.) &78.7 & 69.2 & 86.0 & 80.0 &- & -& -&-\\
NOMAE~\cite{abdelsamad2025nomae}  (fin.)& \textbf{81.8} & \underline{71.0} & \underline{86.5} & 80.8 &71.6 &\underline{59.5} & 76.4& \underline{68.9}\\
Sonata~\cite{wu2025sonata} (fin.) & 79.6 & 70.0  & 86.4 & 80.5 & 69.6 & 58.2& \underline{78.5}&67.7\\
DOS~\cite{abdelsamad2026dos} (fin.)  & \underline{81.5} & 70.5  &  86.0& \underline{81.1}  & \textbf{73.1} &59.2 & 76.3 &68.6\\
\rowcolor{green!8} \coolname{} (fin.)  & 81.1&  \textbf{72.3}& \textbf{87.4}& \textbf{82.3}  & \underline{72.7} & \textbf{60.5} & \textbf{79.6}& \textbf{69.2}\\
\bottomrule
\end{tabular}
}
\caption{Comparison of self-supervised methods on the outdoor scene datasets: nuScenes~\cite{fong2022panoptic} and SemanticKITTI~\cite{behley2019semantickitti}.
}
\label{tab:main_results_outdoor}
\vspace{-2mm}
\end{table}
\PAR{Comparative Study}
Tab.~\ref{tab:main_results} compares \coolname{} with recent self-supervised learning (SSL) methods across three indoor benchmarks: ScanNet~\cite{dai2017scannet}, its long-tailed variant ScanNet200~\cite{rozenberszki2022scannet200}, and S3DIS~\cite{landrieu2018s3dis}. Overall, \coolname{} consistently outperforms prior state-of-the-art SSL frameworks across all datasets and evaluation settings. Remarkably, our method achieves substantial improvements on all three metrics of instance segmentation. On ScanNet and S3DIS, with only linear probing, \coolname{} attains 80–90\% of the supervised performance, highlighting the strong generalization of the learned representations on instance-level tasks.\\
To further validate the generalization of our method, we benchmark on two large-scale outdoor scene datasets: nuScenes~\cite{fong2022panoptic} and SemanticKITTI~\cite{behley2019semantickitti}. We evaluate on the panoptic segmentation task, where both semantic and instance-level predictions jointly contribute to the final score. As shown in Tab.~\ref{tab:main_results_outdoor}, \coolname{} consistently outperforms existing SSL baselines, achieving notable gains of +4.8 PQ on nuScenes and +3.2 PQ on SemanticKITTI. In addition to the improvements in instance-level segmentation, our method also preserves strong semantic segmentation performance, remaining competitive with state-of-the-art approaches. These results demonstrate that \coolname{} not only enhances indoor scene understanding but also generalizes robustly to complex outdoor environments.
\PAR{Component Design}
We conduct an ablation study to validate the effectiveness of each component in our method, as shown in Tab. 4. Simply adding a new branch for offset learning yields only marginal gains. Although applying either regularization strategy alone leads to noticeable improvements, their individual effects remain limited. When the two strategies are combined, however, their complementary strengths result in a larger performance gain, achieving a +3.4\% mAP on instance and +4.8\% PQ increase on panoptic segmentation.\\
\begin{table}[h]
\vspace{-2mm}
\centering
\renewcommand{\arraystretch}{1}
\label{tab:supervised}
\resizebox{0.9\linewidth}{!}{
\begin{tabular}{c|c|c?c|c|c?c|c|c}
\toprule
\multicolumn{3}{c?}{\textbf{Component}} & 
\multicolumn{3}{c?}{\textbf{InsSeg.}}   & \multicolumn{3}{c}{\textbf{PanSeg.}}\\
\midrule
   +$L_{\text{off}}$ & ODR & SCR & mAP & AP$_{50}$ & AP$_{25}$ & PQ &SQ &RQ \\
\midrule
     &  & &28.7 &  49.8& 68.7 & 57.4&82.8 &68.5 \\ 
  \checkmark & & &28.9  &49.6 & 69.1 & 58.5 & 82.3& 69.7   \\ 
\checkmark & \checkmark&  &30.2 & 52.1 & 70.8  &60.4 & 83.6& 71.5\\ 
   \checkmark&  &  \checkmark &  30.5& 51.9 & 71.0& 60.1 & 83.1 & 71.2\\ 
\checkmark&  \checkmark&\checkmark & 32.1&  55.2 & 73.6& 62.2& 84.5& 72.8\\ 
\bottomrule
\end{tabular}}
\caption{Ablation study on \coolname{} components. Models are evaluated on ScanNet~\cite{dai2017scannet} for instance segmentation and nuScenes~\cite{fong2022panoptic} for panoptic segmentation via linear probing.}
\label{tab:ablation_full}
\vspace{-4mm}
\end{table}
\begin{figure*}[t]
\vspace{-3mm}
\centering
    \begin{subfigure}{0.23\linewidth}
    \centering
        \includegraphics[width=\linewidth]{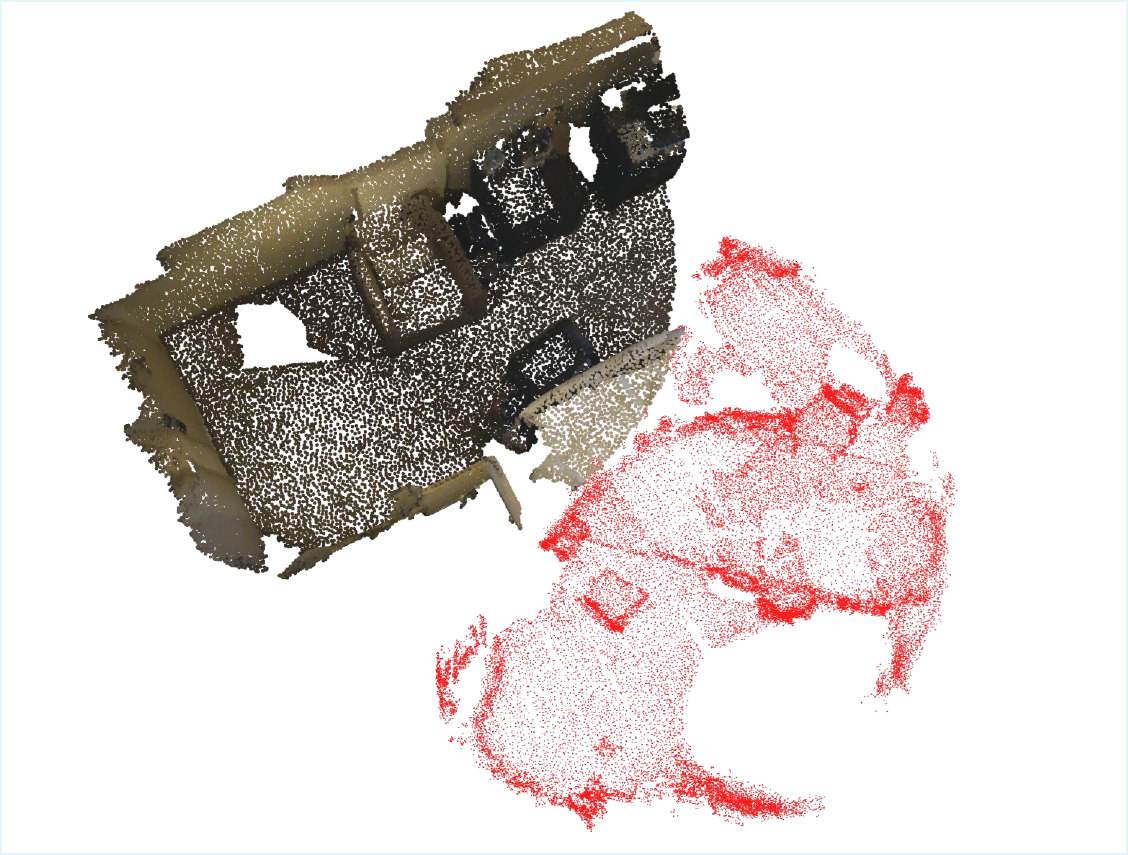}
        \caption{w/o Regularization}
        \label{subfig:without_regularization}
    \end{subfigure}
    \begin{subfigure}{0.23\linewidth}
    \centering
        \includegraphics[width=\linewidth]{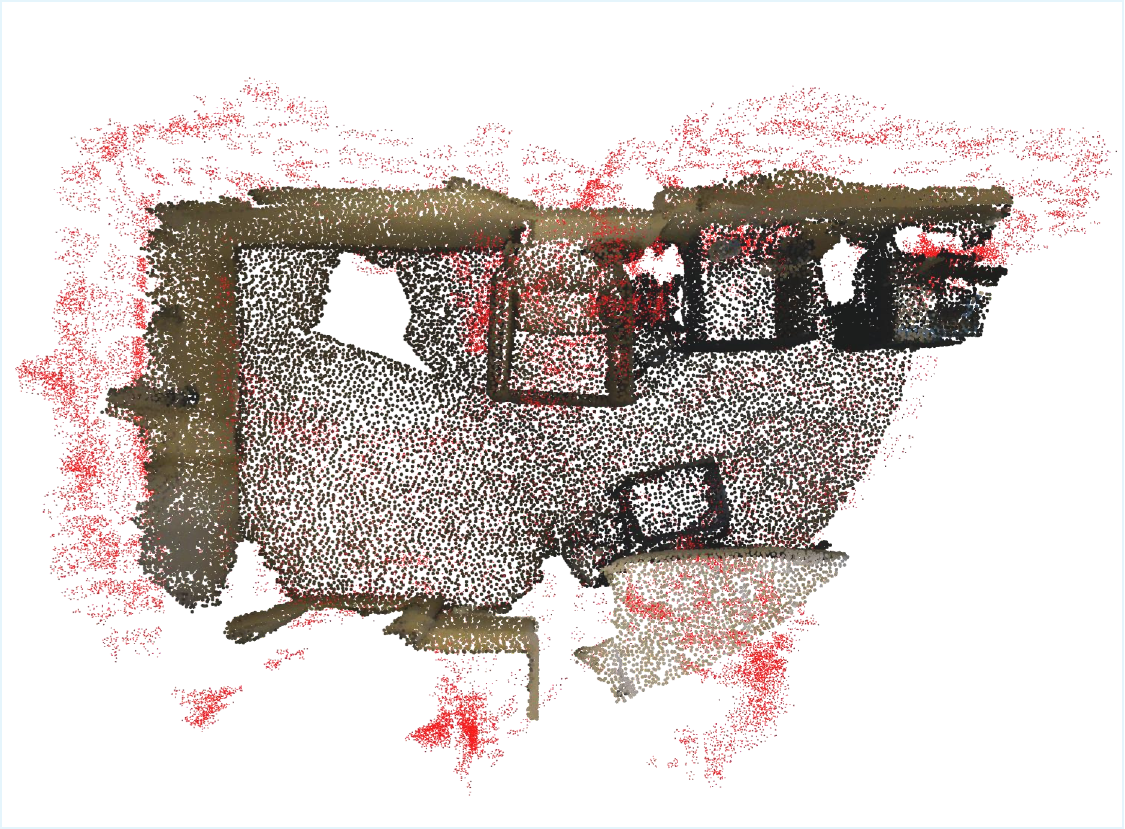}
        \caption{ODR only}
    \end{subfigure}
    \begin{subfigure}{0.23\linewidth}
    \centering
        \includegraphics[width=\linewidth]{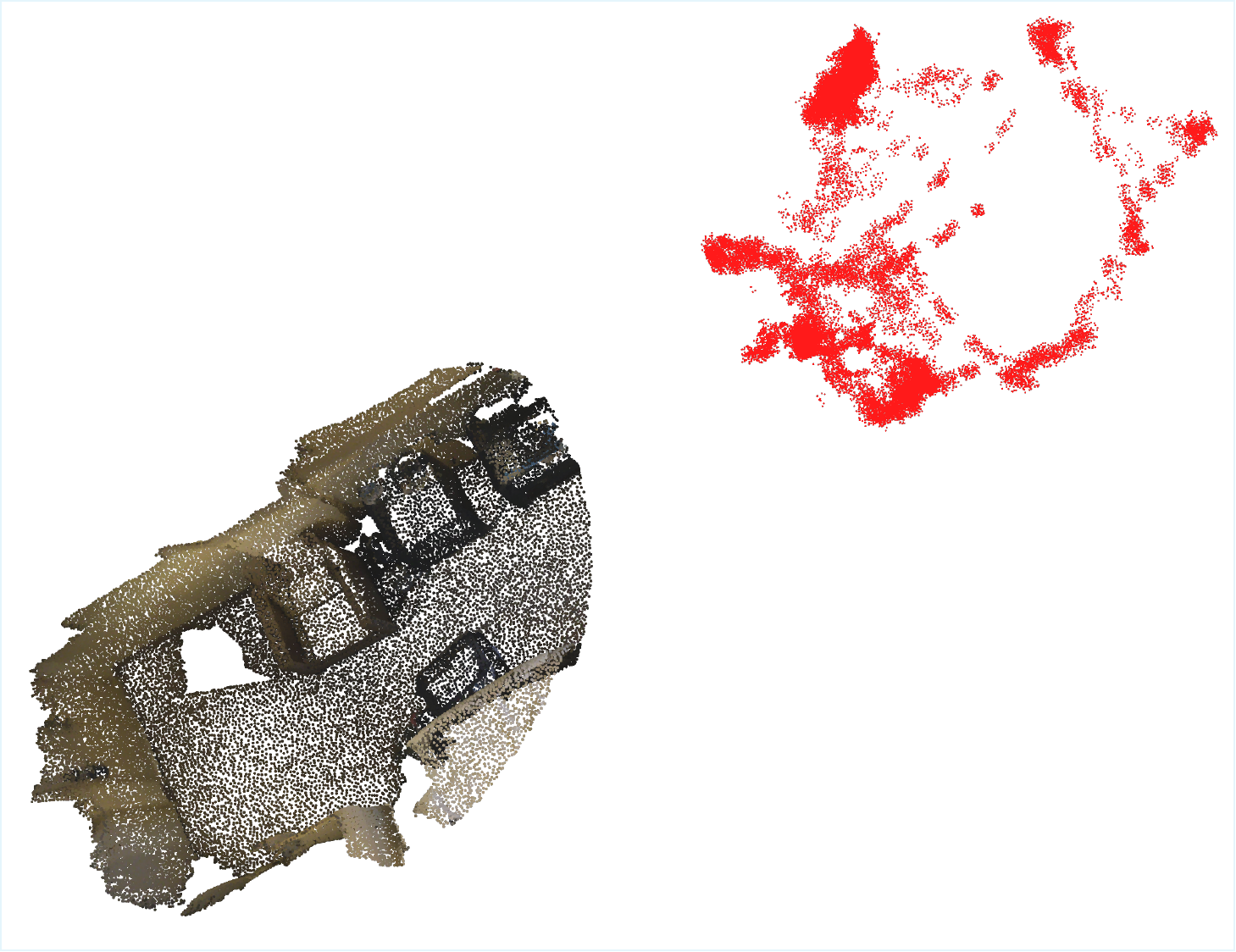}
        \caption{SCR only}
    \end{subfigure}
    \begin{subfigure}{0.23\linewidth}
    \centering
         \includegraphics[width=\linewidth]{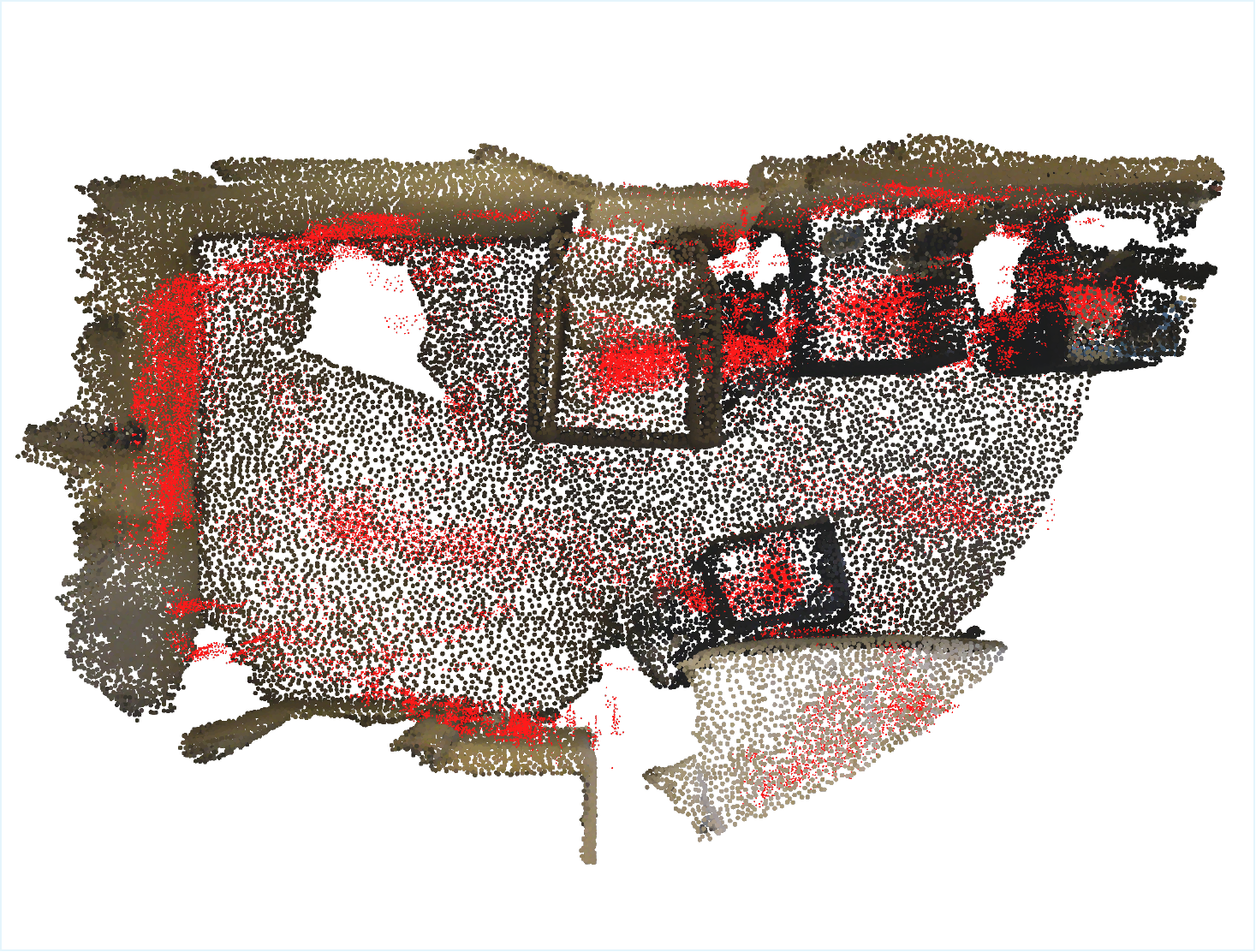}
        \caption{ODR + SCR}
    \end{subfigure}
\caption{Visualization of \textcolor{red}{predicted centroids $c_i = x_i + \mathcal{O}_i$} (\textbf{no explicit regularization}) from models pre-trained under four configurations. (a) Without regularization, the centroids are scattered irregularly across the scenes. (b) With ODR only, the centroids are aligned more closely with the scene structure but they are not grouped (lack of local coherence). (c) With SCR only, the centroids are partially grouped but spatially scattered. (d) With ODR and SCR, the centroids become tightly concentrated around instances.}
\label{fig:regularization_qualitative}
\vspace{-2mm}
\end{figure*}
To further assess how these regularization strategies enhance geometric reasoning during pre-training, we visualize the predicted centroids in Fig.~\ref{fig:regularization_qualitative}. The results show that the model trained with both ODR and SCR produces offsets that are spatially coherent and semantically meaningful, demonstrating a stronger capability for geometric reasoning. These observations qualitatively confirm that our regularization strategies effectively support instance-aware representation learning. Additional ablation studies on key hyperparameters are provided in the Appendix.
\begin{table*}[t]
\vspace{-2mm}
\centering
\resizebox{0.95\linewidth}{!}{
\begin{tabular}{l?c|ccc?c|ccc?c|ccc?c|ccc}
\toprule
\multirow{3}{*}{\textbf{Method}} & 
\multicolumn{4}{c?}{\textbf{ScanNet val}} &  \multicolumn{4}{c?}{\textbf{S3DIS Area5}} & \multicolumn{4}{c?}{\textbf{nuScenes val}}  & 
\multicolumn{4}{c}{\textbf{SemanticKITTI val}} \\
\cmidrule(lr){2-5} \cmidrule(lr){6-9} \cmidrule(lr){10-13} \cmidrule(lr){14-17} 
&SemSeg.& \multicolumn{3}{c?}{InsSeg.} &SemSeg.& \multicolumn{3}{c?}{InsSeg.}&SemSeg.& \multicolumn{3}{c?}{PanSeg.} &SemSeg.& \multicolumn{3}{c}{PanSeg.}\\
\cmidrule(lr){2-2} \cmidrule(lr){3-5} \cmidrule(lr){6-6} \cmidrule(lr){7-9} \cmidrule(lr){10-10} \cmidrule(lr){11-13} \cmidrule(lr){14-14} \cmidrule(lr){15-17}
& mIoU & mAP & AP$_{50}$ & AP$_{25}$ & mIoU & mAP & AP$_{50}$ & AP$_{25}$ & mIoU &PQ& SQ & RQ& mIoU &PQ& SQ & RQ\\ 
\midrule
\rowcolor{gray!7} PTv3 (sup.) & 77.6 & 40.9 & 61.7 & 77.5 & 73.4 & 40.2 &52.1 &61.2 & 80.3 & 69.9 & 86.3 & 80.5  &69.1 & 58.2& 78.7 & 67.7\\
\midrule
Sonata$^\star$~\cite{wu2025sonata} (lin.) & 72.5 & 30.7 & 53.9 & \underline{72.6}& \underline{72.3}& 26.1& 36.6& 45.8 & 66.1 &54.9 & 81.0& 66.0& 62.0 & 50.8 & 76.5 &61.1 \\
DOS$^\star$~\cite{abdelsamad2026dos} (lin.) & \textbf{73.9} & \underline{32.5} & \underline{54.6} & 70.9 & 71.7 &\underline{30.1}& \underline{40.6} & \underline{50.4}   & \textbf{74.8}& \underline{60.9}& \underline{82.6} & \underline{71.5}& \textbf{68.3}& \underline{52.6} &\underline{77.1} &\underline{63.2}\\
\rowcolor{green!8} \coolname{}$^\star$ (lin.) &  \underline{73.5}& \textbf{34.6} & \textbf{57.6} & \textbf{76.3} & \textbf{72.6}& \textbf{37.0} & \textbf{49.8} & \textbf{58.6}  & \underline{74.6} & \textbf{64.7} & \textbf{84.9} &\textbf{74.2} & \underline{68.0} & \textbf{55.0} & \textbf{78.0}& \textbf{65.2}\\
\bottomrule
\end{tabular}
}
\caption{Comparison of self-supervised methods with additional data for pre-training.
}
\vspace{-2mm}
\label{tab:main_results_scalup}
\end{table*}

 
\PAR{Sensitivity of ODR} Offset Distribution Regularization (ODR) requires a prior distribution of instance offsets from annotated datasets, particularly their magnitudes. Since our framework focuses on self-supervised learning, we have to assume no access to labels during training. To verify the effectiveness of ODR in such scenario, we evaluate ODR’s sensitivity by testing various distribution priors, including those derived from real datasets. As shown in Tab.~\ref{tab:odr_sensitivity}, our method exhibits strong robustness to distribution choice. Even when the prior is from the dataset with totally different scene layout (outdoor$\xleftrightarrow{}$indoor) or it deviates significantly from the empirical one, the performance drop remains minimal while still outperforming the baseline without regularization. To further validate our self-supervised claim, we conduct an experiment in which \coolname{} is trained using offset priors estimated by clustering the 3D input points with HDBSCAN, requiring no annotations whatsoever. Performance remains consistent, confirming that an effective prior does not require fine-grained annotations, instead, a coarse estimate of typical object scales is sufficient. In practice, a long-tailed offset distribution can be reasonably assumed, as points close to instance centroids naturally dominate in 3D point clouds. We also observe that as little as 1\% of annotations suffices to estimate a reliable empirical prior, suggesting that exhaustive labeling is unnecessary for achieving strong performance. To contextualize this finding, we compare against a semi-supervised baseline trained directly with 1\% and 10\% ground-truth labels. Notably, the semi-supervised setting only reaches comparable performance when using approximately 10\% labels, highlighting the practical advantage of our regularization-based approach. 
%
\begin{minipage}{\linewidth}
\vspace{0.5mm}
    \centering
    \includegraphics[width=0.9\linewidth]{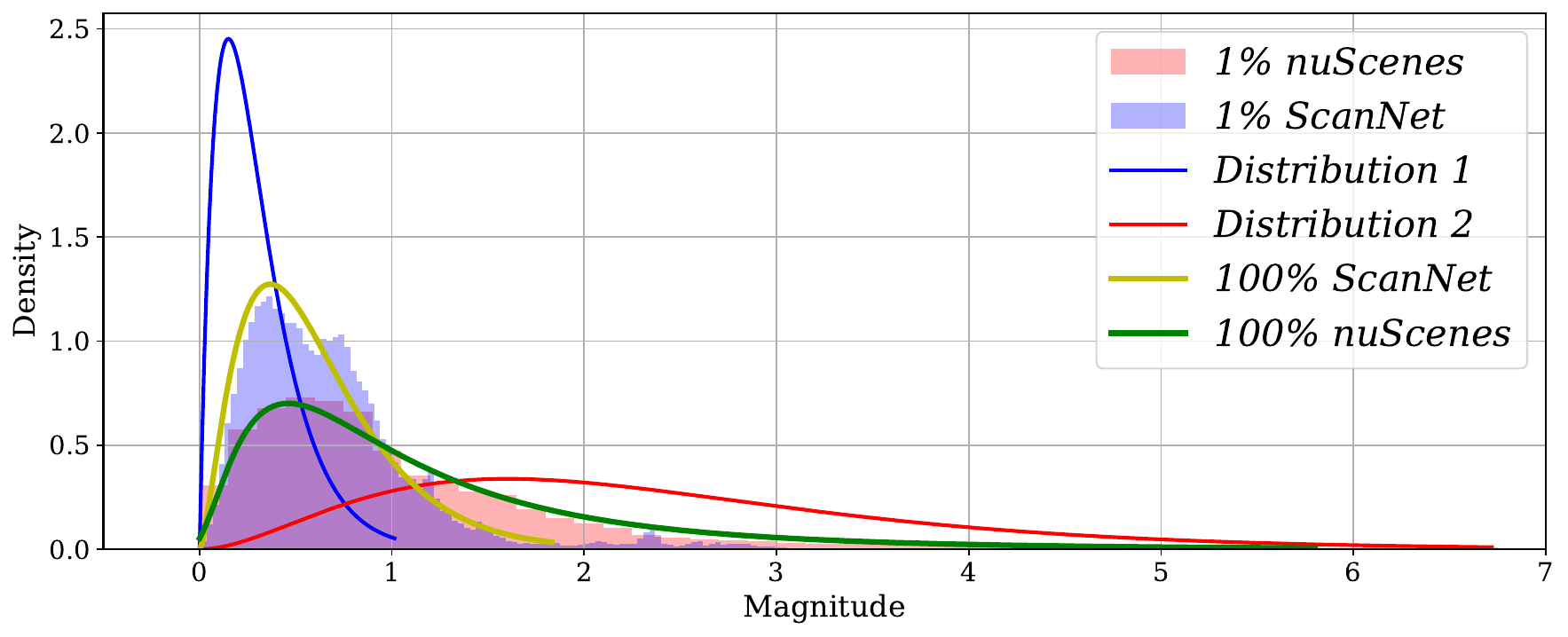}
    \centering
    \resizebox{0.9\linewidth}{!}{
     \begin{tabular}{l?ccc?ccc}
        \toprule
        \multirow{2}{*}{\textbf{Dist. Type}} & 
        \multicolumn{3}{c?}{\textbf{ScanNet val}} & 
        \multicolumn{3}{c}{\textbf{nuScenes val}} \\
        \cmidrule(lr){2-4} \cmidrule(lr){5-7} 
        & mAP & AP$_{50}$ & AP$_{25}$ & PQ&SQ&RQ\\
        \midrule
        w/o Regularization & 28.9 &49.6 & 69.1 & 57.8 &82.8 &68.7  \\
        \midrule
        Dist. 1 & 31.2 & 53.6& 74.1& 60.8 &83.4 &72.2 \\
        Dist. 2& 31.7& 54.2& 72.8 & 61.2 & 83.9 & 72.1\\
       \cellcolor{yellow!10} \textit{ScanNet}& \cellcolor{yellow!10} 32.1& \cellcolor{yellow!10}55.2& \cellcolor{yellow!10}73.6& 62.0& 84.3 &72.5 \\
        \cellcolor{teal!10} \textit{nuScenes}& 31.3& 54.9 & 72.7 &\cellcolor{teal!10}62.2 &\cellcolor{teal!10}84.5 &\cellcolor{teal!10}72.8 \\
        \midrule
        $^\star$HDBSCAN & 31.8& 54.7 &73.1 & 62.1 &84.1 & 72.3 \\
        \bottomrule
        \multicolumn{7}{c}{\textbf{Semi-supervised w/o regularization}}\\
        \midrule
        +1\% labels  & 29.8 & 50.9 & 70.1 &59.3 &83.3 &70.0\\
        +10\% labels & 32.3 & 55.2 & 74.1 & 62.0 & 84.2 & 72.1\\
        \bottomrule
    \end{tabular}
    }
    \captionof{table}{Sensitivity of ODR to different distribution priors applied to offset magnitudes. \textit{ScanNet} and \textit{nuScenes} represent empirical distributions fitted from respective ground-truth instance annotations. Histograms are from 1\% of annotations. $^\star$: Distributions are fitted from unsupervised clustering of 3D point clouds.}
    \label{tab:odr_sensitivity}
\end{minipage}
\begin{table}[t]
\vspace{-2mm}
\centering
\resizebox{\linewidth}{!}{
\begin{tabular}{l|c|ccc}
\toprule
\multirow{2}{*}{\textbf{Method}}& \textbf{SemSeg.} & \multicolumn{3}{c}{\textbf{InsSeg.}} \\
\cmidrule(lr){2-2} \cmidrule(lr){3-5}
 & mIoU & mAP & AP$_{50}$ & AP$_{25}$\\
\midrule
PSA~\cite{nisar2025psa} + \coolname{} & 47.5 {\scriptsize \textcolor{teal}{(+4.6)}}& 14.2 {\scriptsize \textcolor{teal}{(+4.5)}} & 30.8 {\scriptsize \textcolor{teal}{(+10.4)}} & 50.0 {\scriptsize \textcolor{teal}{(+8.1)}}\\
Sonata~\cite{wu2025sonata} + \coolname{}  & 68.6 {\scriptsize \textcolor{teal}{(+1.2)}} & 28.4 {\scriptsize \textcolor{teal}{(+3.4)}} & 50.3{\scriptsize \textcolor{teal}{(+4.2)}} & 67.1 {\scriptsize \textcolor{teal}{(+2.5)}} \\
\bottomrule
\end{tabular}
}
\caption{Evaluation of other SSL frameworks with integration of \coolname{}. Models are evaluated on ScanNet val set under linear probing setting. \textcolor{teal}{(\text{+$\Delta$})} denotes the performance gain.
}
\vspace{-3mm}
\label{tab:generalization_framework}
\end{table}
\PAR{Framework Compatibility}
Our method can be seamlessly integrated into other teacher–student-based frameworks. To evaluate its generalization, we apply \coolname{} to two additional SSL frameworks and report the results in Tab.~\ref{tab:generalization_framework}. Notably, although these baselines like SONATA~\cite{wu2025sonata} already perform strongly, \coolname{} not only enhances instance segmentation but also brings consistent improvements in semantic segmentation performance. Such improvements indicate that semantic and instance objectives are inherently complementary rather than conflicting, and further highlights the broad compatibility and adaptability of our approach within the self-supervised learning paradigm for instance-aware representation learning.
\PAR{Label Efficiency} We evaluate label-efficient training on nuScenes~\cite{fong2022panoptic} under extremely low annotation regimes (0.1\% and 1\% labels), as shown in Tab.~\ref{tab:data_efficiency}. \coolname{} consistently outperforms all self-supervised baselines across all metrics of panoptic segmentation. With only 0.1\% labeled data, \coolname{} achieves 34.9\% in PQ, surpassing supervised performance over 10\%. When the label ratio increases to 1\ it further improves to 42.5\%. These results demonstrate that the instance-aware representations learned by \coolname{} are highly transferable, enabling strong downstream performance even with minimal supervision.
\begin{table}[b]
\vspace{-2mm}
    \centering
    {\fontsize{8}{10}\selectfont
    \begin{tabular}{l?ccc?ccc}
        \toprule
        \multirow{2}{*}{\textbf{Method}} & 
        \multicolumn{3}{c?}{\textbf{0.1\% labels}} & 
        \multicolumn{3}{c}{\textbf{1\% labels}} \\
        \cmidrule(lr){2-4} \cmidrule(lr){5-7} 
        & PQ & SQ & RQ & PQ&SQ&RQ\\
        \midrule
        \rowcolor{gray!7} PTv3 (sup.) & 24.3& 69.0& 30.7& 34.2&71.2 &40.1 \\
        \midrule
        NOMAE~\cite{abdelsamad2025nomae} (fin.) & 30.4 &74.9 & 40.9 & 37.4 & 78.5& 45.2\\
        Sonata~\cite{wu2025sonata} (fin.)& 27.1& 71.4& 34.0& 34.7& 72.5& 42.3\\
        DOS (fin.)& 33.6 & 74.3 & 41.2& 41.2 & 75.7 & 49.2 \\
    \rowcolor{green!8} \coolname{} (fin.)& \textbf{34.9} & \textbf{75.2}& \textbf{42.1}& \textbf{42.5} &\textbf{82.0} & \textbf{49.6}\\
        \bottomrule
    \end{tabular}
    }
    \caption{Label efficient training on nuScenes panoptic segmentation~\cite{fong2022panoptic}.}
    \label{tab:data_efficiency}
    \vspace{-1mm}
\end{table}

\PAR{Layout of Regularization}
We further investigate how different regularization layouts affect performance. As shown in Tab.~\ref{tab:design choice}, applying SCR before ODR leads to a clear drop in instance segmentation accuracy. This suggests that global regularization (ODR) is crucial for stabilizing offset predictions before enforcing local coherence among points. We also explore applying ODR on the student side, where the regularization directly influences gradient updates in the student backbone. This configuration again results in degraded performance, likely due to conflicting gradients and unstable optimization. In contrast, regularizing the teacher outputs provides a stable and structured supervisory signal, allowing the student to adapt smoothly without disrupting representation learning~\cite{caron2021dino, wu2025sonata, oquab2024dinov2}.
\begin{table}[t]
\vspace{-2mm}
\centering
{\fontsize{8}{10}\selectfont
\begin{tabular}{l|c|ccc}
\toprule
\multirow{2}{*}{\textbf{Design}}& \textbf{SemSeg.} & \multicolumn{3}{c}{\textbf{InsSeg.}} \\
\cmidrule(lr){2-2} \cmidrule(lr){3-5}
 & mIoU & mAP & AP$_{50}$ & AP$_{25}$\\
\midrule
 \coolname{}$^\diamond$ & 72.0 {\scriptsize \textcolor{red}{(-0.4)}}& 30.0 {\scriptsize \textcolor{red}{(-2.1)}} & 50.7 {\scriptsize \textcolor{red}{(-4.5)}}& 69.5 {\scriptsize \textcolor{red}{(-4.1)}}\\
\coolname{}$^\dagger$ & 71.9 {\scriptsize \textcolor{red}{(-0.5)}}& 30.9 {\scriptsize \textcolor{red}{(-1.2)}}& 53.9 {\scriptsize \textcolor{red}{(-1.3)}}& 72.8 {\scriptsize \textcolor{red}{(-0.8)}}\\
\coolname{}  & 72.4 & 32.1& 55.2&73.6\\
\bottomrule
\end{tabular}
}
\caption{Comparison of different regularization layouts. All models are evaluated on ScanNet via linear probing. $^\diamond$: order of ODR and SCR is reversed. $^\dagger$: ODR is added on the student side as a regularization loss via KL divergence.
}
\label{tab:design choice}
\vspace{-4mm}
\end{table}
\PAR{Multi-dataset Pre-training}
To explore the scalability of our framework, we expand the pre-training pool by aggregating multiple datasets. For indoor scenes, we combine ScanNet~\cite{dai2017scannet}, S3DIS~\cite{landrieu2018s3dis}, and Structured3D~\cite{Structured3D}, yielding approximately 24k point clouds (significantly smaller than the 140k scenes used by Sonata$^\star$). For outdoor scenes, we follow prior protocols~\cite{abdelsamad2026dos, wu2025sonata} and mix nuScenes~\cite{fong2022panoptic}, SemanticKITTI~\cite{behley2019semantickitti}, and Waymo~\cite{mei2022waymo} during pre-training. The results in Tab.~\ref{tab:main_results_scalup} show that performance consistently improves as the scale of pre-training data increases. Remarkably, under this scale-up setting, \coolname{} approaches supervised performance under linear probing, particularly on instance and panoptic segmentation. This strong scalability suggests that \coolname{} progresses a promising step toward unified 3D foundation models.

\section{Conclusion}
\label{sec:conclusion}
In this work, we introduce \coolname{}, a novel self-supervised learning framework that enables point cloud encoders to jointly learn semantic invariance and instance-aware geometric reasoning without labels. By introducing two complementary regularization strategies applied to the geometric learning, our method prevents model collapse while enriching feature representations with geometric priors and enforcing local coherence. \coolname{} achieves state-of-the-art performance in instance and panoptic segmentation across five benchmarks. We believe this work marks an important step toward scalable 3D foundation models capable of holistic scene understanding.
\PAR{Limitations and Future Work} While \coolname{} achieves substantial improvements over existing self-supervised methods under the linear probing setting, a noticeable gap still remains compared to fully supervised performance. Scaling up pre-training with more data or jointly using indoor and outdoor datasets could advance further progress. In addition, extending the framework to incorporate spatiotemporal cues for 4D geometric reasoning represents a promising avenue for future exploration.

\clearpage
\section*{Acknowledgements}
We are grateful to Professor Abhinav Valada of the Robot Learning Lab at the University of Freiburg for his invaluable guidance, insightful discussions, and continuous support throughout this work. As the PhD advisor of our second author, Mohamed Abdelsamad, his expertise and thoughtful feedback significantly shaped the direction and quality of this research.

{
    \small
    \bibliographystyle{ieeenat_fullname}
    \bibliography{main}
}
\clearpage
\setcounter{page}{1}
\maketitlesupplementary
\appendix

\section{Additional Implementation details}

\PAR{Details of SCR} Spatial Clustering Regularization (SCR) plays a key role in \coolname{} by enforcing local geometric coherence among points, enabling the model to learn instance-level geometric reasoning. We provide the full procedure of SCR in Algorithm~\ref{alg:scr}. The process consists of two stages: (i) global feature-based grouping and (ii) local spatial refinement. First, we apply K-means clustering to partition points into $K$ coarse semantic groups. This step leverages the strong semantic awareness preserved in self-supervised backbones. Next, for each segment $S_k$, we compute ODR-regularized predicted centroids and construct a $k$-nearest-neighbor graph using two constraints: (1) neighbor count $k_{\mathrm{nn}}$ and (2) maximum neighbor distance $\tau_d$. This pruning ensures that connections only form between spatially consistent points rather than long-range neighbors. We then apply a standard BFS algorithm to decompose the graph into multiple connected components. Components smaller than a minimum size threshold $\tau_{\mathrm{min}}$ are discarded to avoid noisy groupings. Each remaining component is treated as a pseudo-instance, from which a refined centroid is computed. The target offsets are then defined as the displacement from each point to its assigned pseudo-centroid. These targets supervise the student model via self-distillation, promoting stable local centroid alignment rather than random spatial drift.

\PAR{Pre-taining Settings} We summarize all hyperparameter configurations in Tab.~\ref{tab:hyper}. For single-dataset pre-training and downstream fine-tuning, we use a single NVIDIA H200 GPU. For multi-dataset pre-training, we use two H200 GPUs in Distributed Data Parallel (DDP) mode.  Following Sonata~\cite{wu2025sonata} and DOS~\cite{abdelsamad2026dos}, we concatenate multi-scale features from the last three encoder stages rather than using only the final stage.

\begin{table}[t]
\centering
\resizebox{1\linewidth}{!}{
\begin{tabular}{l |c |c}
\toprule
\textbf{Config} & \textbf{Outdoor} & \textbf{Indoor} \\
\midrule
optimizer & AdamW  & AdamW \\
scheduler & Cosine& Cosine \\
learning rate & 2e-3  & 4e-3 \\
weight decay & 4e-2  & 4e-2 \\
batch size & 16  & 16 \\
Datasets & nuScenes / Sem.Kitti  & Scannet / Scannet200 / S3DIS \\
Mask Ratio & 0.7 & 0.7 \\
Mask Size & 1 m & 40 cm \\
warmup ratio & 0.05  & 0.05 \\
training epochs & 50  & 800/800/3000\\
$\alpha_{zipf}$ & 1.3  & 0.1 / 1.3 / 0.1 \\
warmup ratio of $L_{\text{off}}$ & 0.1 & 0.1 \\
$\lambda_{\text{off}}$ & 0.25 & 0.25 \\
$K$ (K-means) & 20 & 20 \\
$iter$ (K-means) & 10 & 10 \\
$k_{nn}$ & 20 & 150\\
$\tau_d$ & 1 m & 120 cm\\
$\tau_{min}$ & 10 & 30\\
 Distribution ($\mathcal{D}$) & Uniform(0, 1) & Uniform(0, 1)  \\
 Distribution ($\mathcal{M}$) & LogNormal($\mu$=0, $\sigma$=0.76) & Gamma($a$=0.24, $\theta$=2.53) \\
\bottomrule
\end{tabular}
}
\caption{Pretraining settings for indoor and outdoor point clouds.}
\vspace{-3mm}
\label{tab:hyper}
\end{table}

\PAR{Multi-Dataset Pre-training} To enhance cross-domain generalization, we pre-train \coolname{} jointly on multiple datasets. For outdoor settings, we follow standard practice and train on the combined corpus of nuScenes~\cite{caesar2020nuscenes}, SemanticKITTI~\cite{behley2019semantickitti}, and Waymo~\cite{mei2022waymo}, totaling approximately 116k point clouds. A single unified model is trained across all datasets using the same architecture and hyperparameters as in single-dataset pre-training. For indoor settings, we pre-train on Structured3D~\cite{Structured3D}, ScanNet~\cite{dai2017scannet}, and S3DIS~\cite{landrieu2018s3dis}, comprising roughly 24k point clouds. Following prior works~\cite{abdelsamad2026dos, wu2025sonata}, we scale up model capacity to better handle the increased structural diversity of indoor environments, expanding the encoder to \([3, 3, 3, 12, 3]\) blocks with channel widths \([48, 96, 192, 384, 512]\). This configuration yields a 108M-parameter model, compared to 38M in the default setup.

\PAR{Indoor Instance Segmentation}
We evaluate \coolname{} on three indoor benchmarks. \textbf{ScanNet}~\cite{dai2017scannet} contains 1,613 RGB-D scans with 3D instance annotations, split into 1,201 training, 312 validation, and 100 test scenes. \textbf{ScanNet200}~\cite{rozenberszki2022scannet200} extends ScanNet with fine-grained labels over 200 semantic categories. Following standard protocol, we report instance segmentation using the 18 canonical instance classes shared with ScanNet. \textbf{S3DIS}~\cite{landrieu2018s3dis} consists of 271 indoor scenes across six areas annotated with 13 semantic classes, all of which are evaluated for instance segmentation. We adopt the common \textit{Area-5} protocol, where Area 5 serves as the test split and the remaining areas for training. We report mean Average Precision (mAP) as the primary metric. $\text{AP}_{25}$ and $\text{AP}_{50}$ denote AP at 25\% and 50\% IoU thresholds, while AP averages scores from 50\% to 95\% IoU (step size 5\%).

\PAR{Outdoor Panoptic Segmentation}
We evaluate on two large-scale LiDAR benchmarks. \textbf{SemanticKITTI}~\cite{behley2019semantickitti} contains 22 driving sequences captured by a 64-beam LiDAR sensor with 19 semantic classes; sequences 00--10 (excluding 08) are used for training, 08 for validation, and 11--21 for testing. \textbf{nuScenes}~\cite{caesar2020nuscenes} includes 1000 urban driving scenes from Boston and Singapore collected with a 32-beam LiDAR sensor. Following~\cite{fong2022panoptic}, we evaluate 16 merged semantic classes. Panoptic segmentation performance is measured using the Panoptic Quality (PQ):
\begin{equation}
\text{PQ} = 
\underbrace{\frac{\sum_{(i,j) \in \text{TP}} \text{IoU}(i,j)}{|\text{TP}|}}_{\text{Segmentation Quality (SQ)}}
\times
\underbrace{\frac{|\text{TP}|}{|\text{TP}| + \frac{1}{2}(|\text{FP}| + |\text{FN}|)}}_{\text{Recognition Quality (RQ)}},
\end{equation}
where SQ measures segmentation accuracy and RQ measures instance recognition. We additionally report SQ and RQ separately.

\section{Additional Experiments}
\begin{figure}[h]
\centering
    \begin{subfigure}{0.48\linewidth}
    \centering
        \includegraphics[width=\linewidth]{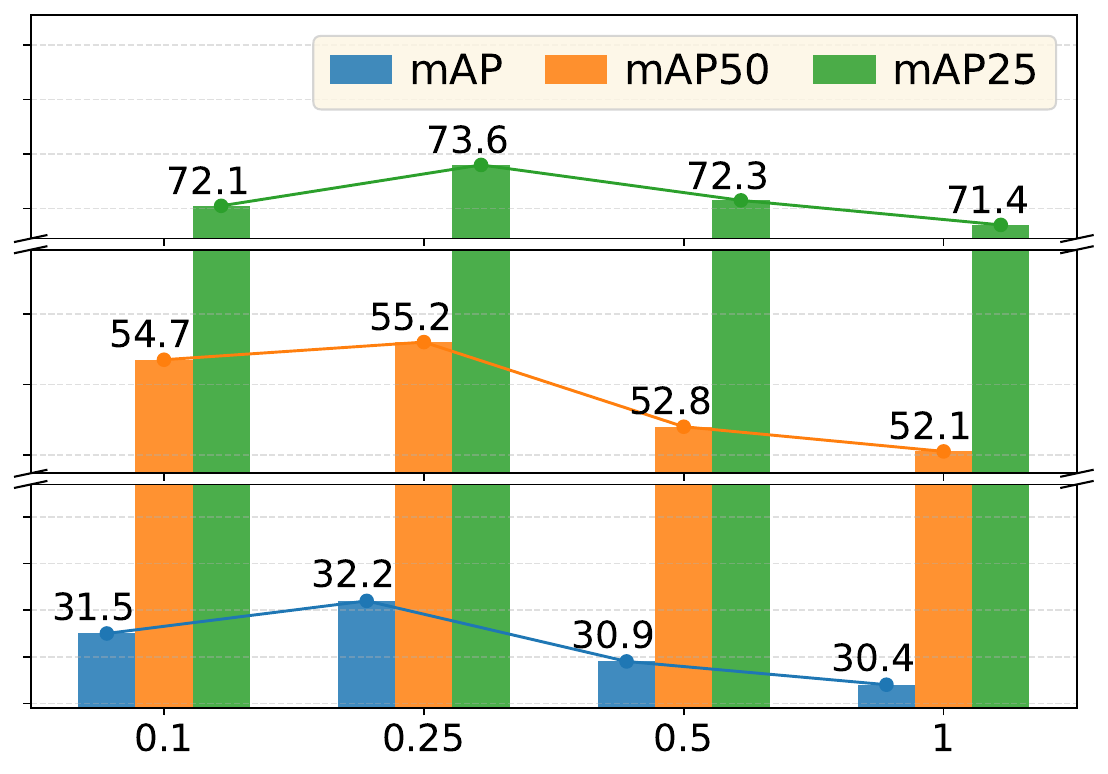}
        \caption{Loss weight $\lambda_{\text{off}}$}
        \label{subfig:loss_weight}
    \end{subfigure}
    \begin{subfigure}{0.48\linewidth}
    \centering
        \includegraphics[width=\linewidth]{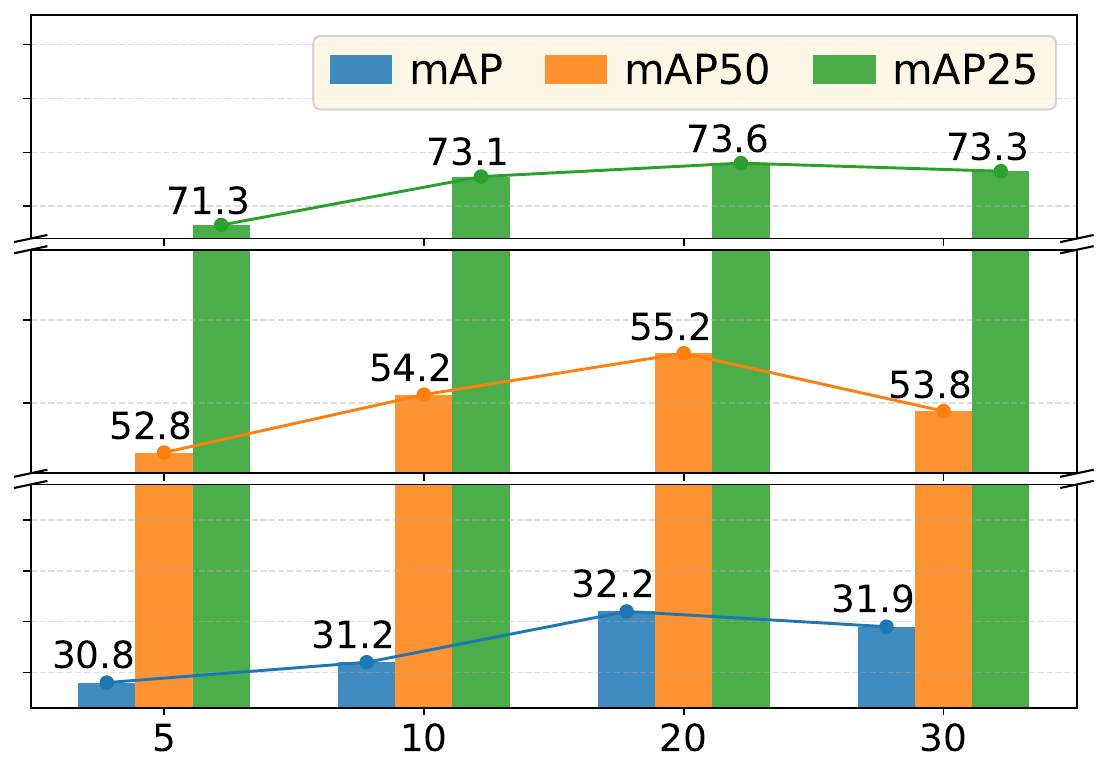}
        \caption{Number of K-Means clusters $K$}
        \label{subfig:num_kmeans}
    \end{subfigure}
\caption{Results of different parameter settings on ScanNet.}
\label{fig:ablation_params}
\vspace{-3mm}
\end{figure}

\PAR{Effect of $\lambda_{\text{off}}$ and $K$} We study the impact of two key parameters in our method: the offset loss weight $\lambda_{\text{off}}$ and the number of K-means clusters $K$ used in SCR. The results are shown in Fig.~\ref{fig:ablation_params}. Regarding the loss weight, $\lambda_{\text{off}}=0.25$ achieves the optimal performance among all tested values. For K-means clustering, using $K=20$ yields the best performance, which intuitively aligns with the semantic category distribution of the dataset. More broadly, $K$ controls a fundamental trade-off: too small value risks merging semantically distinct but feature-similar regions, while too large value tends to over-segment individual objects into spurious parts. Empirically, we found moderate values of $K$ is sufficient, as most scenes contain a limited number of distinct semantic concepts.
\PAR{Effect of Warmup Ratio of $L_{\text{off}}$} In Tab.~\ref{tab:ablation_offset_warmup}, we study how different warmup ratios for the offset loss affect instance segmentation performance on ScanNet. Compared to no warmup, introducing a short warmup phase significantly improves results, with the best performance observed at a ratio of 0.1. This suggests that the model benefits from first establishing stable semantic representations before learning offsets. However, increasing the warmup duration beyond 0.1 leads to a gradual performance drop, as excessive delay reduces the effective training time for geometric reasoning. Overall, these results highlight the importance of gradually introducing the offset loss to stabilize early optimization while still allowing sufficient training for instance awareness.
\begin{table}[h]
\centering
{\fontsize{9}{11}\selectfont
\begin{tabular}{c|ccc}
\toprule
\textbf{Warmup Ratio} & mAP & AP$_{50}$ & AP$_{25}$\\
\midrule
0.0 & 30.8& 53.7& 72.8\\
0.1 & 32.1& 55.2& 73.6\\
0.2 & 31.7 &  54.6& 73.1\\
0.4 & 31.4 &53.5 &72.4 \\
0.6 & 30.1 & 52.6 &71.6 \\
0.8 & 28.8 & 49.9& 68.1\\
\bottomrule
\end{tabular}
}
\caption{Results of different warmup ratio of $L_{\text{off}}$ on ScanNet
}
\vspace{-4mm}
\label{tab:ablation_offset_warmup}
\end{table}

\PAR{Runtime Analysis} Since \coolname{} builds upon DOS~\cite{abdelsamad2026dos}, the additional offset branch and two regularization steps introduce extra computation during pre-training. Overall, the total pre-training time increases by approximately 25\% (e.g. from 20 to 25 hours on ScanNet~\cite{dai2017scannet} and from 24 to 29 hours on nuScenes~\cite{caesar2020nuscenes}), measured on a single GPU.

\PAR{Object Detection} To further verify the generalization of our approach, we evaluate \coolname{} on the nuScenes object detection benchmark~\cite{caesar2020nuscenes} using CenterPoint~\cite{yin2021center} as the detector, with results reported in Tab.~\ref{tab:object_detection}. Under decoder probing, the pre-trained encoder is frozen and only the remaining detector components are trained. Under finetuning, all model weights are optimized end-to-end. In both settings, \coolname{} outperforms existing SSL approaches by a significant margin, demonstrating its transferability across diverse downstream tasks and representing a promising step toward holistic 3D foundational perception.

\begin{table}[h]
    \centering
    {\fontsize{8}{10}\selectfont
    \begin{tabular}{l?cc?cc}
        \toprule
        \multirow{2}{*}{\textbf{Method}} & \multicolumn{2}{c?}{\textbf{OD Prob.}} &  \multicolumn{2}{c}{\textbf{OD 1\%}}  \\
        \cmidrule(lr){2-3}\cmidrule(lr){4-5} 
        &mAP&NDS &mAP&NDS\\
        \midrule
        Sonata~\cite{wu2025sonata} &44.6 &55.0 &41.3 & 52.8 \\
        DOS~\cite{abdelsamad2026dos} & 55.4 & 61.5& 49.0& 58.0 \\
        \coolname{} & \textbf{56.7}& \textbf{62.5}& \textbf{50.8}& \textbf{60.2} \\
        \bottomrule
    \end{tabular}
    }
    \caption{OD Prob.: Object detection under decoder probing protocol. OD 1\%: Object detection finetuning on 1\% annotations.}
    \label{tab:object_detection}
    \vspace{-3mm}
\end{table}

\PAR{Cross-dataset Probing}
Beyond single- and multi-dataset pre-training, we further evaluate \coolname{} under a cross-dataset probing setting, where the model is pre-trained on one dataset and linearly probed on another. As shown in Tab.~\ref{tab:cross_dataset_probing}, \coolname{} consistently outperforms existing SSL baselines across both transfer directions, confirming that the instance-aware representations learned by \coolname{} generalize robustly across different outdoor scene layouts.
\begin{table}[h]
    \centering
    {\fontsize{8}{10}\selectfont
    \begin{tabular}{l?ccc?ccc}
        \toprule
        \multirow{2}{*}{\textbf{Method}} &
        \multicolumn{3}{c?}{\textbf{SK$\xrightarrow{}$Nu}} & 
        \multicolumn{3}{c}{\textbf{Wa$\xrightarrow{}$Nu}} \\
         \cmidrule(lr){2-4} \cmidrule(lr){5-7} 
        & PQ & SQ & RQ & PQ&SQ&RQ\\
        \midrule
        Sonata~\cite{wu2025sonata} & 31.0 & 72.5 & 40.5 & 36.9& 75.4& 47.1 \\
        DOS~\cite{abdelsamad2026dos}  & 46.4 & 78.9& 57.5&51.4 & 80.3& 62.6 \\
        \coolname{} & \textbf{56.7} & \textbf{80.1} & \textbf{59.4}& \textbf{54.8}& \textbf{81.9} & \textbf{66.1}\\
        \bottomrule
    \end{tabular}
    }
    \caption{Results on panoptic segmentation under cross-dataset probing setting. SK: SemanticKITTI~\cite{behley2019semantickitti}, Nu: nuScenes~\cite{caesar2020nuscenes}, Wa: Waymo Open Dataset~\cite{mei2022waymo}}
    \vspace{-4mm}
    \label{tab:cross_dataset_probing}
\end{table}
\begin{algorithm}[h]
\caption{Spatial Clustering Regularization (SCR)}
\label{alg:scr}
\begin{algorithmic}[1]
\Require Teacher features $\mathbf{F} = \{f_i\}_{i=1}^N$, coordinates $\{x_i\}_{i=1}^N$, ODR-regularized offsets $\{\tilde{\mathcal{O}}_i\}_{i=1}^N$
\Require Hyperparameters: number of neighbors $k_{nn}$, distance threshold $\tau_d$, minimum instance size $\tau_{min}$
\Ensure Pseudo-instance targets $\{\mathcal{O}^\ast_i\}_{i=1}^N$

\State \textbf{Predict centroids:}
\[
\hat{c}_i \gets x_i + \tilde{\mathcal{O}}_i
\]

\State \textbf{Global feature grouping:}
\[
\{S_1,\dots,S_K\} \gets \mathrm{KMeans}(\mathbf{F}; K, iter)
\]

\State \textbf{Local spatial clustering per segment:} 
\Statex For each segment $S_k$: 
\Statex (a) Compute $k$ nearest neighbors for each point $i \in S_k$ using Euclidean distance in centroid space:
\[
\mathcal{N}(i) = \mathrm{kNN}(\hat{c}_i, k_{nn})
\]
Retain edges only if distance is below threshold:
\[
j \in \mathcal{N}(i) \quad \textbf{iff} \quad \|\hat{c}_i - \hat{c}_j\|_2 < \tau_d
\]

\Statex (b) Extract connected components using Breadth-First Search (BFS):
\[
\mathcal{I}_k = \mathrm{BFS}(\mathcal{N})
\]

\Statex(c) Remove small components (noise filtering):
\[
\mathcal{I}_k \gets \{ I_{k,j} \in \mathcal{I}_k \mid |I_{k,j}| \ge \tau_{min} \}
\]

\State \textbf{Compute refined centroids:}
\[
\bar{c}_{k,j} = \frac{1}{|I_{k,j}|} \sum_{i \in I_{k,j}} \hat{c}_i
\]

\State \textbf{Generate new offset targets:}
\[
\mathcal{O}^\ast_i = \bar{c}_{k,j} - x_i, \quad \forall i \in I_{k,j}
\]

\State \Return $\{\mathcal{O}^\ast_i\}_{i=1}^N$

\end{algorithmic}
\end{algorithm}
\section{Unsupervised Instance Segmentation}
\begin{figure}[b]
    \centering
    \vspace{-5mm}
    \begin{subfigure}[b]{0.23\textwidth}
        \centering
        \includegraphics[width=\linewidth]{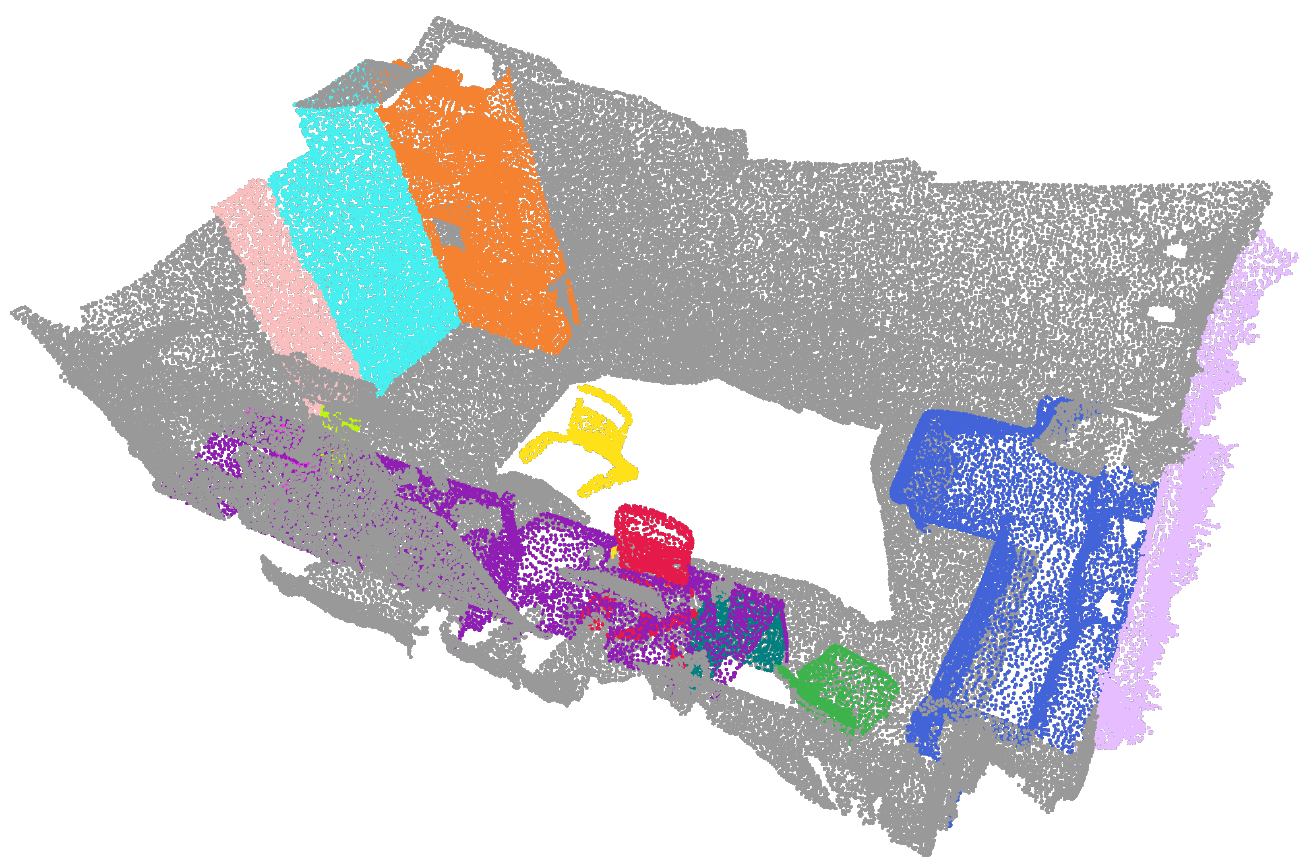}
        \label{fig:row1_a}
    \end{subfigure}
    \hfill
    \begin{subfigure}[b]{0.23\textwidth}
        \centering
        \includegraphics[width=\linewidth]{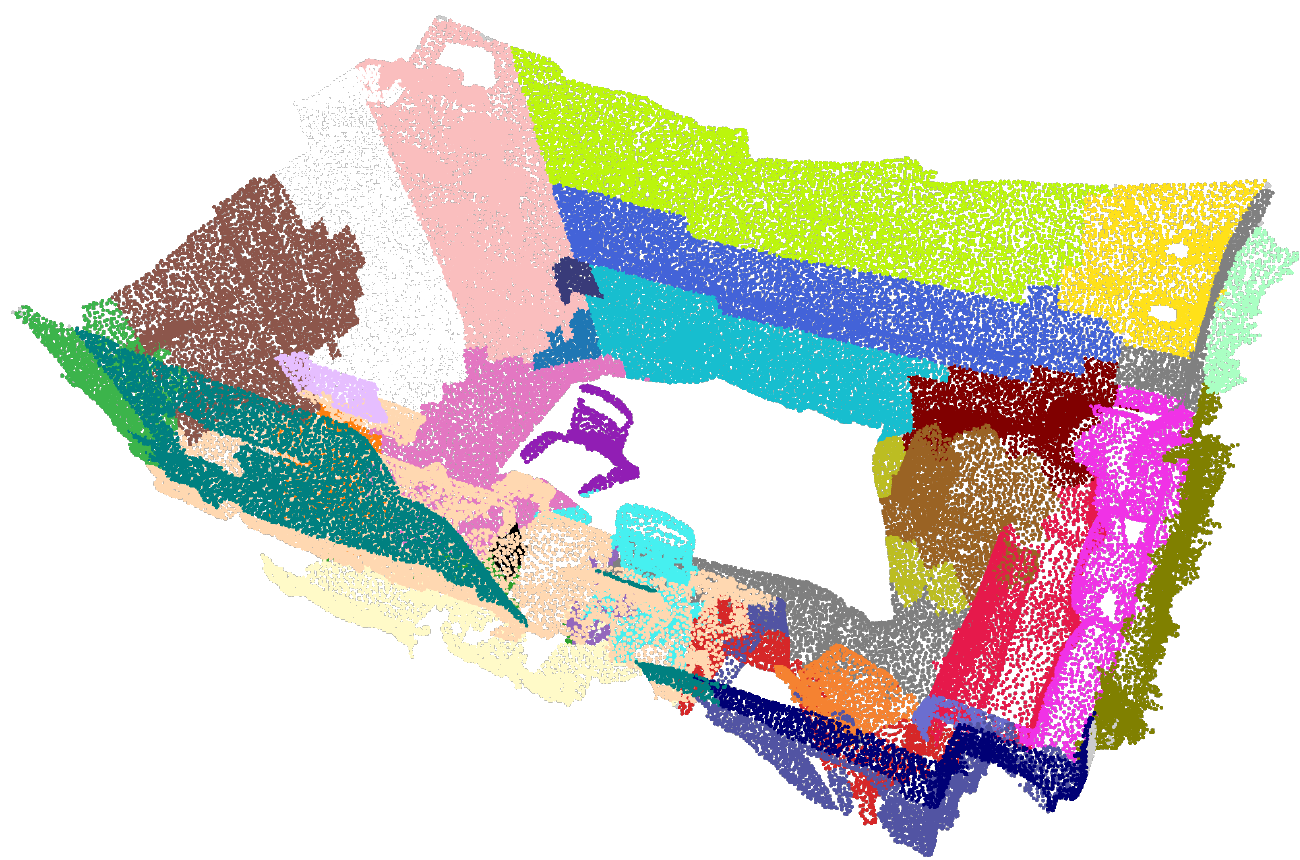}
        \label{fig:row1_b}
    \end{subfigure}
    \hfill
    \begin{subfigure}[b]{0.23\textwidth}
        \centering
        \includegraphics[width=\linewidth]{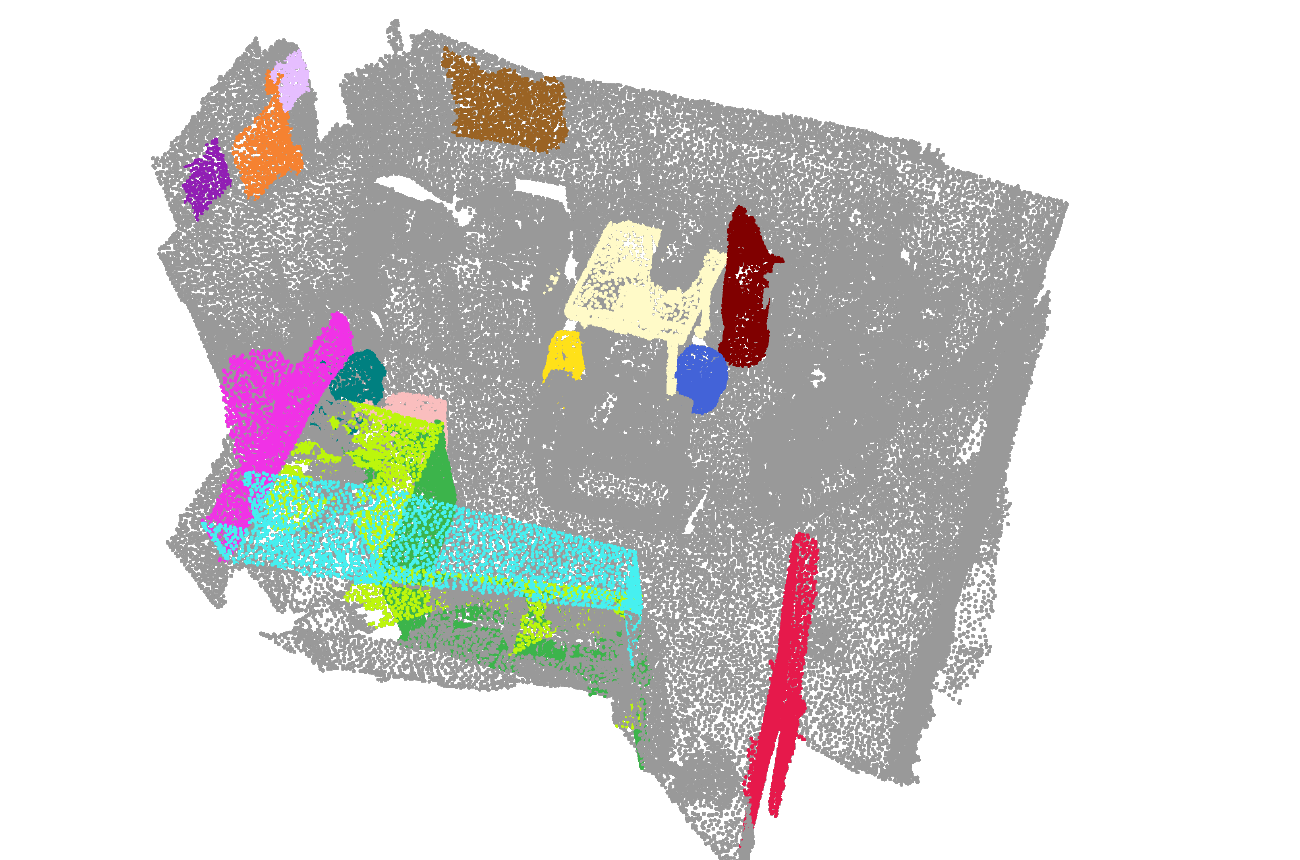}
        \label{fig:row2_a}
    \end{subfigure}
    \hfill
    \begin{subfigure}[b]{0.23\textwidth}
        \centering
        \includegraphics[width=\linewidth]{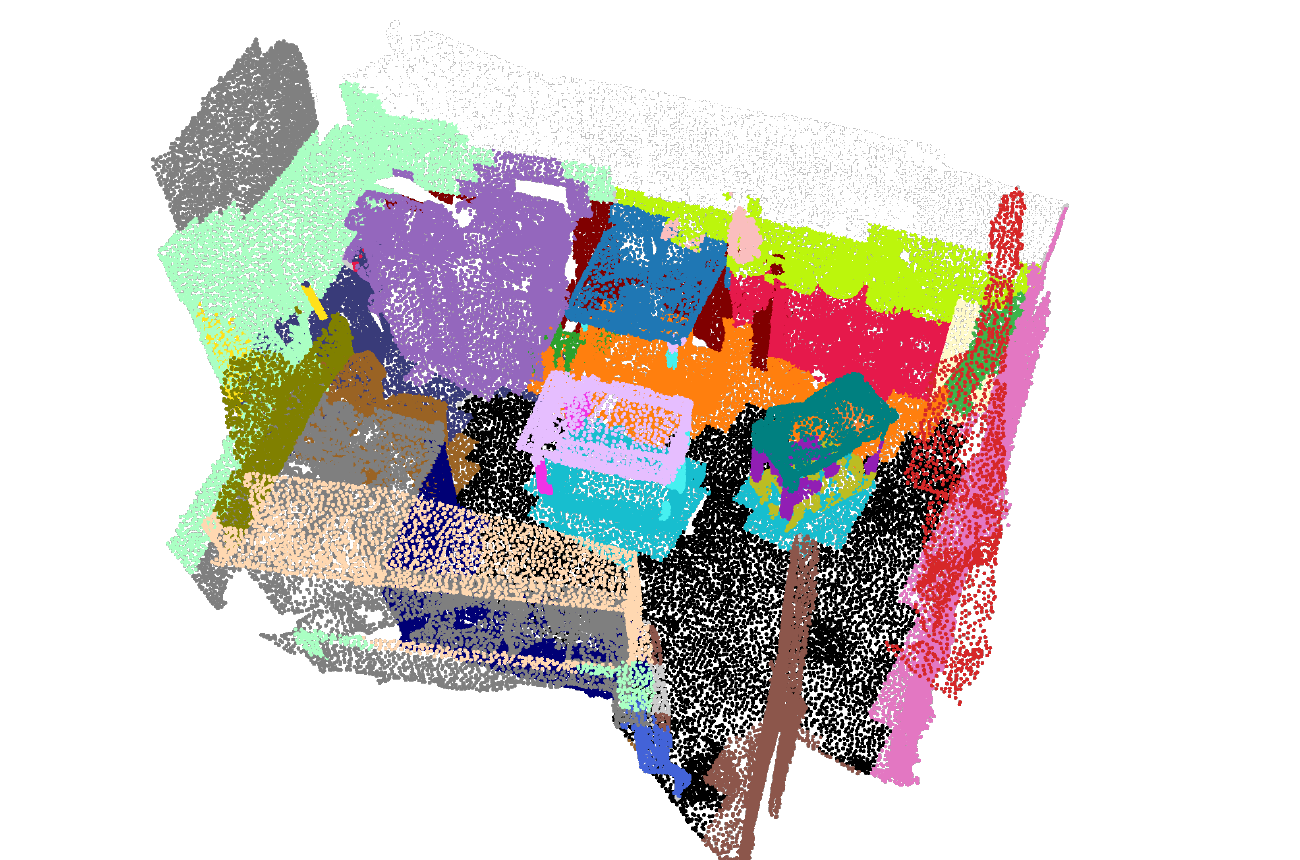}
        \label{fig:row2_b}
    \end{subfigure}

    \hfill
    \begin{subfigure}[b]{0.23\textwidth}
        \centering
        \includegraphics[width=\linewidth]{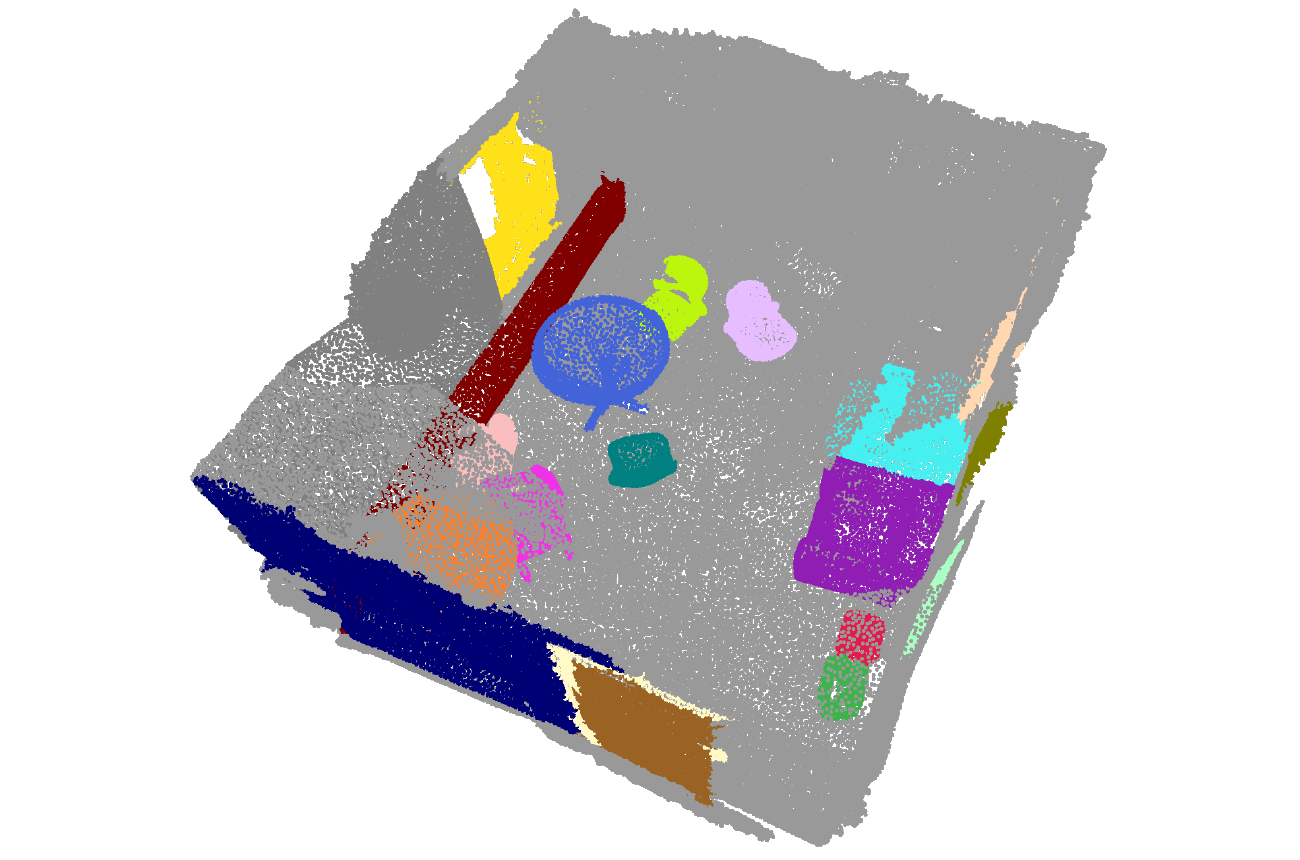}
        \label{fig:row3_a}
    \end{subfigure}
    \hfill
    \begin{subfigure}[b]{0.23\textwidth}
        \centering
        \includegraphics[width=\linewidth]{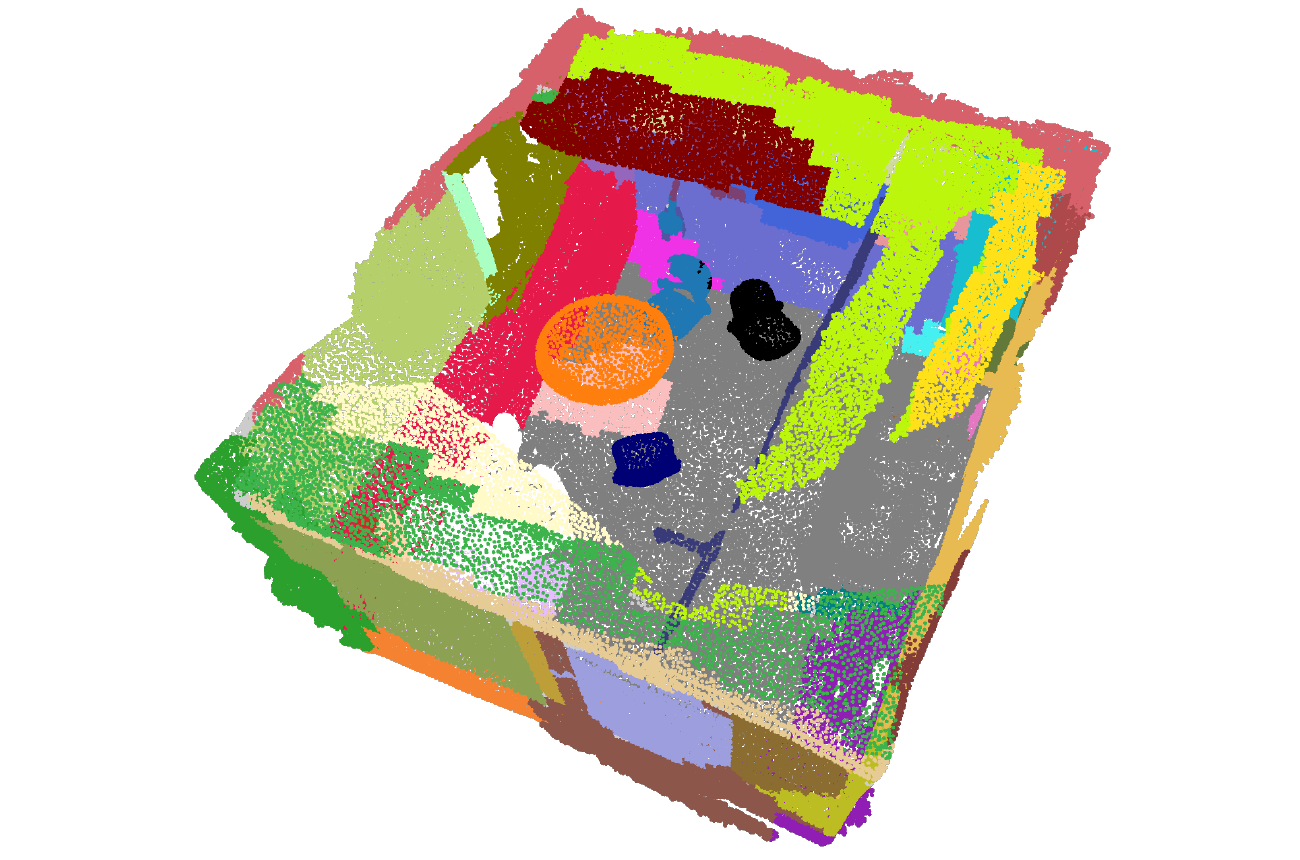}
        \label{fig:row3_b}
    \end{subfigure}
    \begin{subfigure}[b]{0.23\textwidth}
        \centering
        \includegraphics[width=\linewidth]{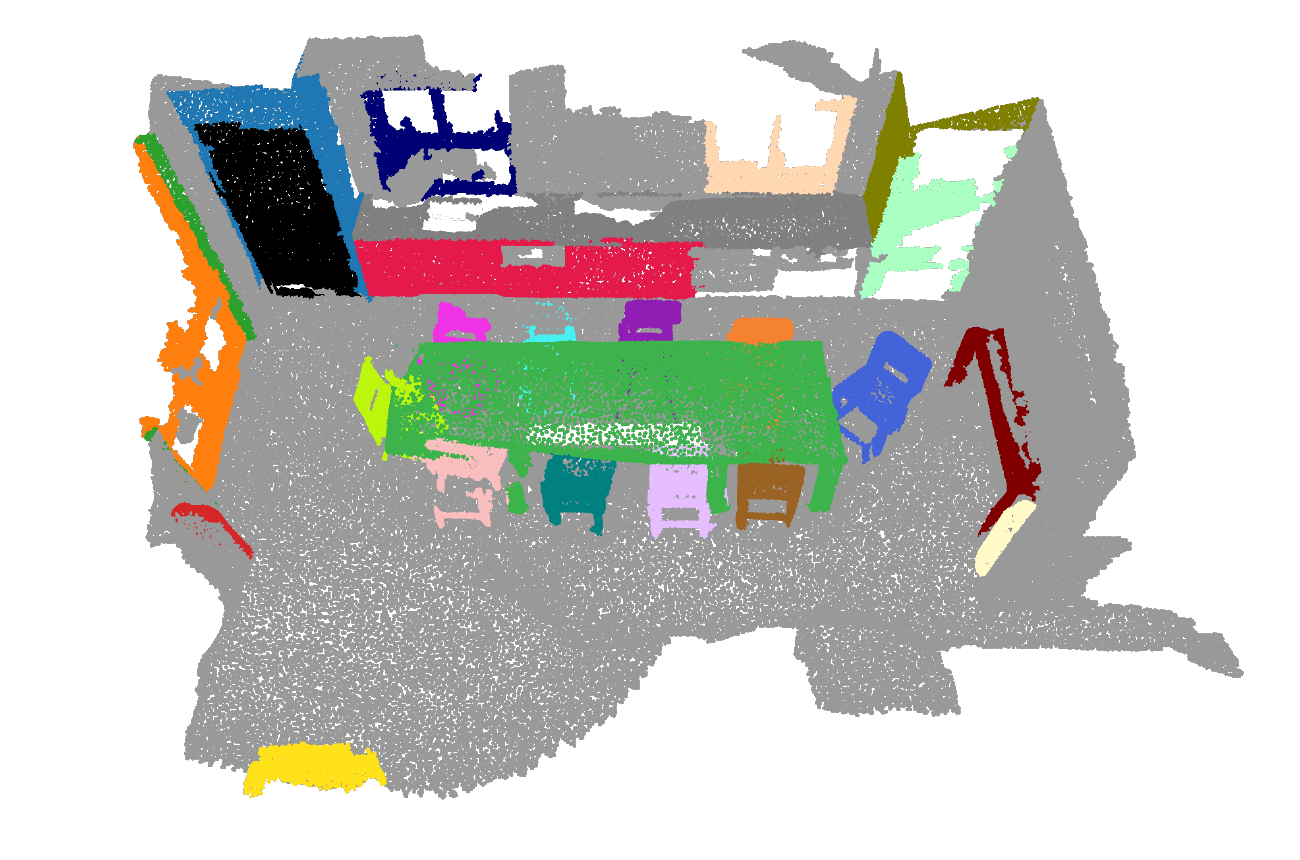}
        \label{fig:row4_a}
    \end{subfigure}
    \hfill
    \begin{subfigure}[b]{0.23\textwidth}
        \centering
        \includegraphics[width=\linewidth]{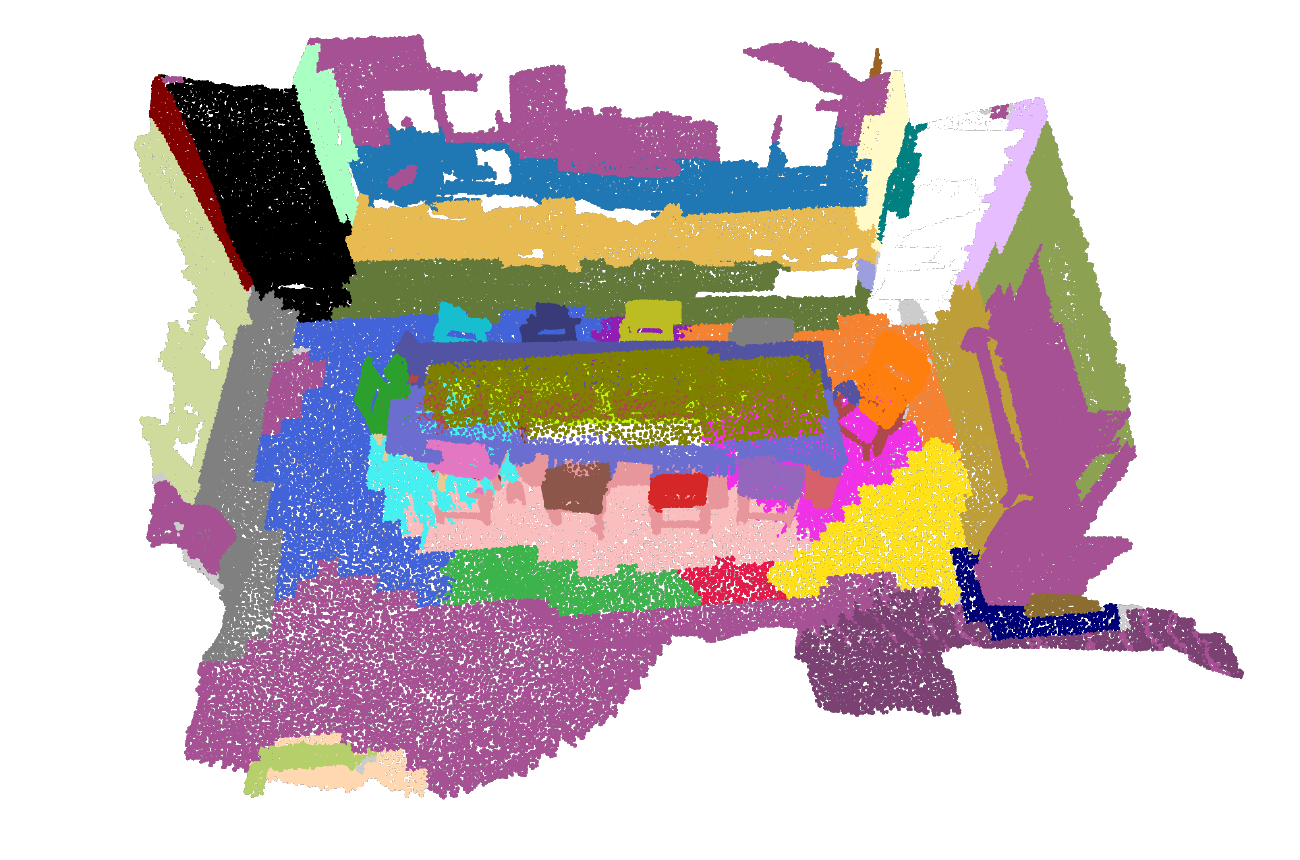}
        \label{fig:row4_b}
    \end{subfigure}
    \begin{subfigure}[b]{0.23\textwidth}
        \centering
        \includegraphics[width=\linewidth]{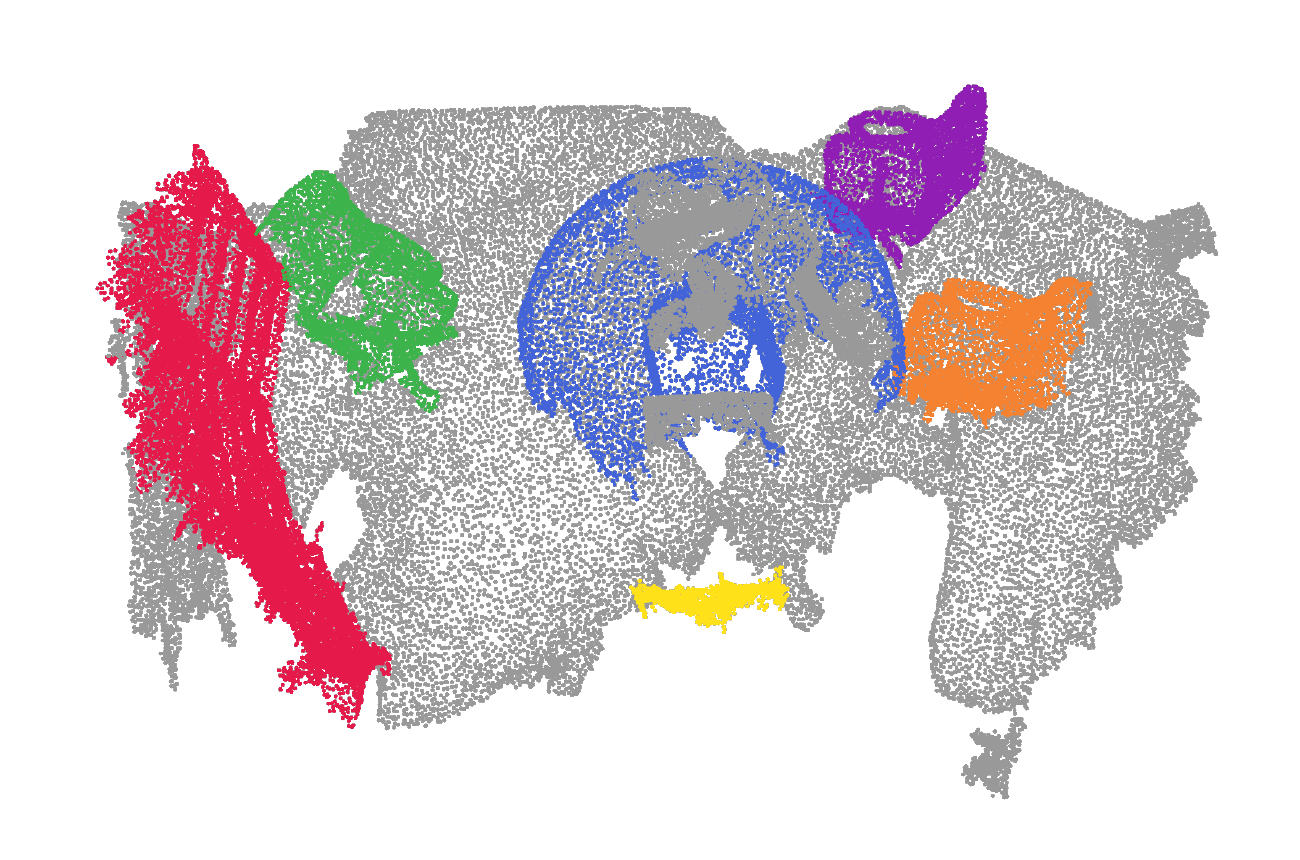}
        \label{fig:row4_a}
    \end{subfigure}
    \hfill
    \begin{subfigure}[b]{0.23\textwidth}
        \centering
        \includegraphics[width=\linewidth]{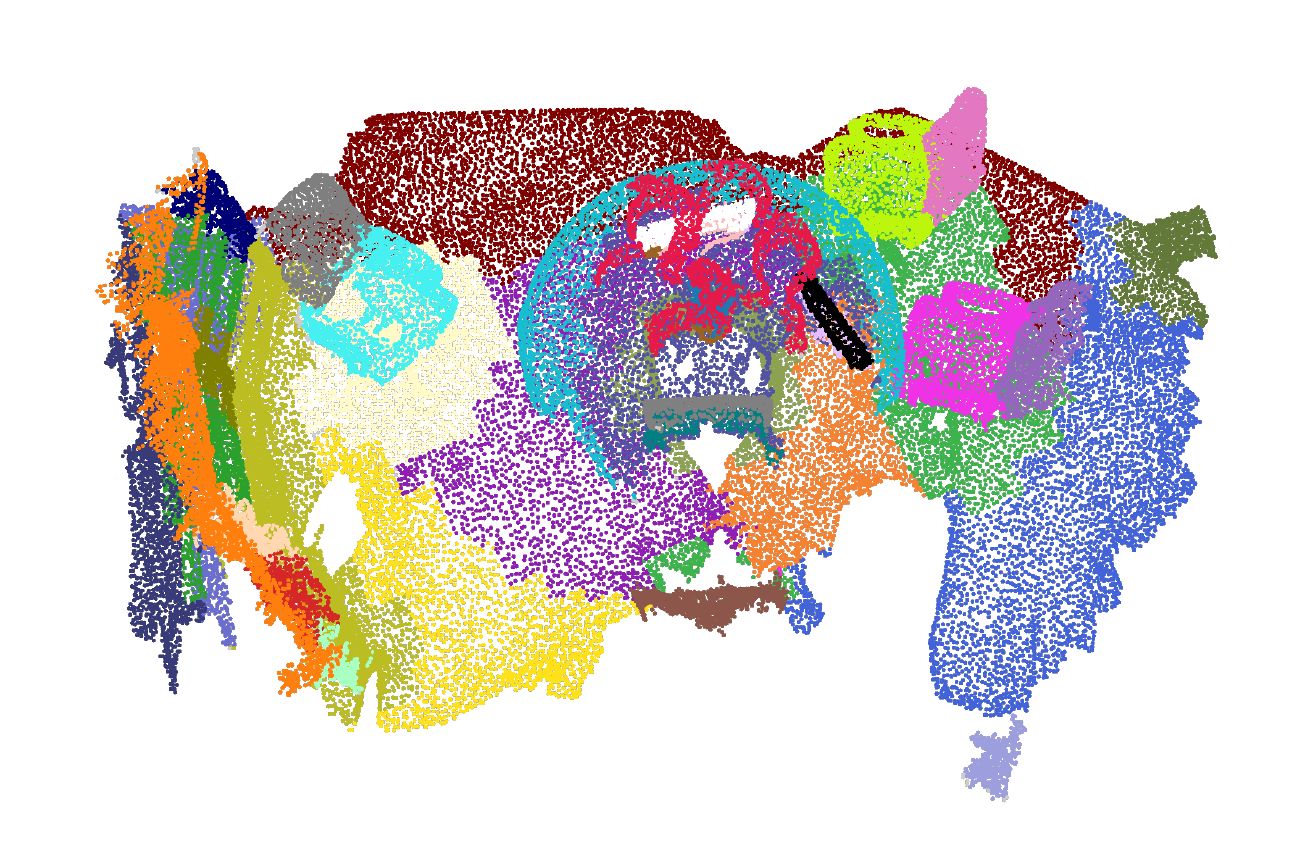}
        \label{fig:row4_b}
    \end{subfigure}
    \caption{Qualitative results of unsupervised instance segmentation. Left: ground-truth instance labels. Right: predictions from \coolname{} obtained directly from offset clustering without any downstream supervision.}
    \label{fig:qualitative_unsup}
    \vspace{-3mm}
\end{figure}

We further assess whether \coolname{} produces useful instance-aware representations without any downstream supervision. Instead of training an instance segmentation head, we directly use the offsets predicted by the pretrained model and perform BFS-based clustering on the offset-shifted centroids to generate instance proposals. The resulting clusters are matched to ground-truth instances via Hungarian assignment for evaluation. As shown in Tab.~\ref{tab:unsupervised}, \coolname{} substantially outperforms classical unsupervised baselines such as HDBSCAN~\cite{hdbscan} and Felzenszwalb clustering~\cite{felzenszwalb}, demonstrating strong geometric reasoning and robust instance separation ability even without task-specific optimization. These results highlight the broader utility of our approach beyond standard self-supervised settings. In addition to quantitative results, we present qualitative evaluations of unsupervised instance segmentation in Fig.~\ref{fig:qualitative_unsup}.
\begin{table}[h]
\centering
{\fontsize{8}{10}\selectfont
\begin{tabular}{c|ccc}
\toprule
\textbf{Method} & mAP & AP$_{50}$ & AP$_{25}$\\
\midrule
HDBSCAN~\cite{hdbscan} & 1.7& 4.2& 16.4\\
 Felzenszwalb~\cite{felzenszwalb}& 1.2 & 2.3& 13.4 \\
  \coolname{}& 10.8& 18.7& 43.4\\
\bottomrule
\end{tabular}
}
\caption{Results of unsupervised instance segmentation on ScanNet.
}
\label{tab:unsupervised}
\vspace{-4mm}
\end{table}




\end{document}